\documentclass[11pt,letterpaper]{article}
\date{}
\usepackage{subfigure}
\usepackage{subcaption}
\usepackage[authoryear]{natbib}
\usepackage[letterpaper, total={6.5in, 9.5in}]{geometry}

\usepackage{hyperref}
\usepackage{url}
\usepackage{booktabs}
\usepackage{tabularx}
\usepackage{graphicx}
\usepackage{amsmath}
\usepackage{amssymb}
\usepackage{mathrsfs}
\usepackage[scr=rsfso]{mathalpha}
\usepackage{xcolor}
\newtheorem{example}{Example}[section]

\title{Gone With the Bits: Revealing Racial Bias in Low-Rate Neural Compression for Facial Images}

\author{Tian Qiu\thanks{Equal contribution} ,~~Arjun Nichani\footnotemark[1] ,~~Rasta Tadayontahmasebi,~~Haewon Jeong\\
University of California, Santa Barbara\\
\texttt{\{tian\_qiu,anichani,rasta,haewon\}@ucsb.edu}
}

\begin{document}
\maketitle

\begin{abstract}
    Neural compression methods are gaining popularity due to their superior rate-distortion performance over traditional methods, even at extremely low bitrates below 0.1 bpp. As deep learning architectures, these models are prone to bias during the training process, potentially leading to unfair outcomes for individuals in different groups. In this paper, we present a general, structured, scalable framework for evaluating bias in neural image compression models. Using this framework, we investigate racial bias in neural compression algorithms by analyzing nine popular models and their variants. Through this investigation, we first demonstrate that traditional distortion metrics are ineffective in capturing bias in neural compression models. Next, we highlight that racial bias is present in all neural compression models and can be captured by examining facial phenotype degradation in image reconstructions. We then examine the relationship between bias and realism in the decoded images and demonstrate a trade-off across models. Finally, we show that utilizing a racially balanced training set can reduce bias but is not a sufficient bias mitigation strategy. We additionally show the bias can be attributed to compression model bias and classification model bias. We believe that this work is a first step towards evaluating and eliminating bias in neural image compression models.
\end{abstract}

\section{Introduction}\label{sec:intro}

Lossy image compression aims to accurately represent images using a minimal number of bits while maintaining their perceptual quality in reconstructions. This area has been the focus of extensive research for the past 40 years, and image encoders/decoders (``codecs") such as JPEG~\citep{wallace1991jpeg}, BPG~\citep{bpg}, and even the latest hand-engineered codec in VVC~\citep{bross2021overview} have been crucial enabling technologies in the modern digital world. Despite the widespread adoption in everyday use, traditional codecs are insufficient for extreme scenarios with low-bandwidth availability, such as space~\citep{gao2023extremely}, underwater~\citep{li2023rfd}, low-power communication systems~\citep{ez2018analysis} and low-latency systems~\citep{hu2021joint}. These extreme scenarios impose a very narrow information bottleneck that limits the reconstruction quality of traditional codecs. In recent years, neural network-based compression (``neural compression'') has emerged as a popular compression method that enables image compression under extremely low-bitrate scenarios. Early works in this field \citep{toderici2015variable, toderici2017full} utilize recurrent neural networks, while many subsequent studies have employed VAE-based architectures~\citep{balle2018variational, townsend2019practical, duan2023qarv, duan2023lossy}. Recent studies explore leveraging modern generative architectures such as GANs \citep{agustsson2019generative, mentzer2020high} and Diffusion \citep{yang2023lossy} to promote higher levels of realism in reconstructions. 

The goal of this paper is to examine potential unwanted biases in low-rate neural compression models. We consider a scenario where we train a neural compression model, specialized for human faces, to attain a very low bitrate. Regardless of the compression method used, image reconstructions at low bitrates will inherently suffer from significant distortion due to the insufficient number of bits used to represent images.  The central question we pose is the following: \textit{when we train a neural network models to compress human faces with low bitrates, would the model degrade facial images equally across different demographic groups? Or, would it prioritize accurately reconstructing one racial group's faces, at the expense of sacrificing image qualities of another racial group when the information bandwidth is limited?} 
Such biased and unfair performance  of neural compression can have a significant impact on people of marginalized groups, especially in extreme and high-risk scenarios where low-rate compression schemes are deployed (e.g., delaying rescue operations due to inaccurate facial images transmitted in a warzone).

This question is inspired by a line of research that studies related questions. 
In \citet{yucer2022does}, the authors investigate bias in face image compression using the traditional JPEG scheme and show unequal performance in facial recognition tasks across different racial groups. Recent works~\citep{jalal2021fairness, laszkiewicz2024benchmarking,Tanjim_2022_BMVC} also looked at biases of image construction using  
neural networks. 
Although these works differ from our setting in that they start with downsampled or heavily corrupted facial images and use neural networks only for denoising or super-resolution, we see a fundamental connection to our work: downsampling or adding noise can be viewed as imposing a narrow information bottleneck, similar to compression. In these settings, it was shown that the reconstructed images often show a specific type of distortion---African American faces are frequently reconstructed to appear more Caucasian, while Caucasian faces largely retain their original features---a phenomenon referred to as the ``White Obama'' problem~\citep{jalal2021fairness,laszkiewicz2024benchmarking}. Despite these works, to the best of knowledge, \emph{our work is the first to examine bias in neural compression models, consisting of a neural network encoder and decoder. }

\begin{figure}[t] 
    \centering
    \begin{minipage}{0.47\textwidth}
        \centering
        \includegraphics[width=\linewidth]{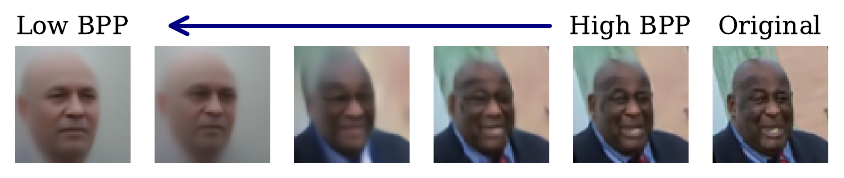}
        \caption*{(a) African}
    \end{minipage}
    \begin{minipage}{0.47\textwidth}
        \centering
        \includegraphics[width=\linewidth]{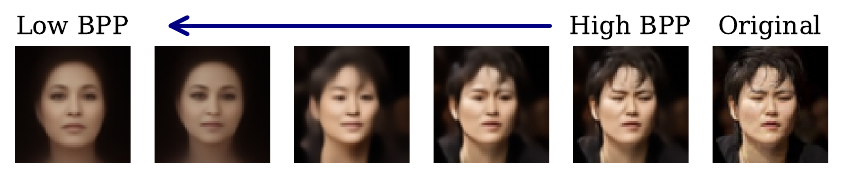}
        \caption*{(b) Asian}
    \end{minipage}
    \begin{minipage}{0.47\textwidth}
        \centering
        \includegraphics[width=\linewidth]{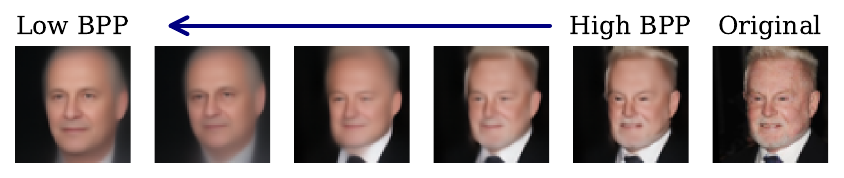}
        \caption*{(c) Caucasian}
    \end{minipage}
    \begin{minipage}{0.47\textwidth}
        \centering
        \includegraphics[width=\linewidth]{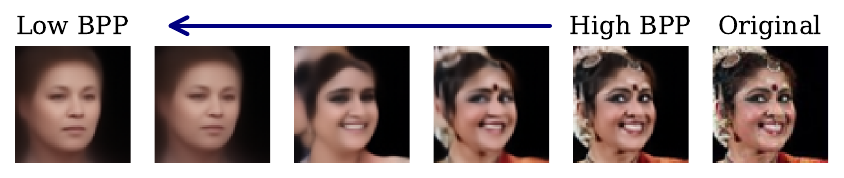}
        \caption*{(d) Indian}
    \end{minipage}
    \caption{All the neural compression models in our evaluation exhibit \textbf{bias in skin type} for \textbf{African racial group}. Examples are from the \textit{QRes}~\citep{duan2023lossy} model. As compression bitrate reduces, African faces gradually experience skin-lightening effects, while other racial groups are impacted less. Our novel evaluation approach with phenotype classifiers quantifies how different phenotypes degrade and highlights bias in this process.}\label{fig:intro_visualization}
\end{figure}

To comprehensively explore our central question, we propose the following research questions:
\textbf{RQ1}. Do neural compression models exhibit bias, and how can we quantify this bias? \textbf{RQ2}. How does bias vary across different model architectures? \textbf{RQ3}. Does using a balanced dataset reduce or eliminate bias? \textbf{RQ4}. Why are neural compression models discriminatory? 
To answer the research questions, we design a general framework and metric to evaluate bias in neural image compression models and benchmark nine state-of-the-art neural compression models, including VAE-, GAN-, and diffusion-based models as summarized in Table~\ref{model_table}. The key observations from our in-depth analysis are:

\begin{table}[t]
  \centering 
    \caption{\textbf{Neural Compression Models.} The evaluated neural compression models and variants vary in network architecture, optimization objectives, and rate control strategies.} 
    \begin{tabular}{c | c c c c }
    \toprule
    Model & Fixed-rate & Architecture &  Realism & Rates [bpp]\\
    \midrule
    Hyperprior~\cite{balle2018variational} & \checkmark & VAE & $\times$ & 0.04 - 0.52 \\
    Joint~\cite{minnen2018joint} & \checkmark & VAE & $\times$      & 0.02 - 0.52  \\  
    GaussianMix-Attn~\cite{cheng2020learned} & \checkmark & VAE & $\times$ & 0.03 - 0.45  \\
    QRes~\cite{duan2023lossy} & $\times$ & VAE & $\times$ & 0.01 - 0.70   \\
    VarQRes~\cite{duan2023qarv} & $\times$ & VAE & $\times$ & 0.10 - 0.51   \\
    HiFiC~\cite{mentzer2020high} & \checkmark & VAE + GAN & $\checkmark$ & 0.04 - 0.52  \\
    CDC~\cite{yang2023lossy} & \checkmark & Diffusion & $\times$ & 0.01 - 0.55  \\
    CDC-L2~\cite{yang2023lossy} & \checkmark & Diffusion & $\times$ & 0.15 - 0.50  \\
    CDC-LPIPS~\cite{yang2023lossy} & \checkmark & Diffusion & $\checkmark$ & 0.07 - 0.52  \\
    \bottomrule
    \end{tabular}%
    \label{model_table}
\end{table}%
 
\begin{itemize}
    \item Traditional image distortion cannot effectively capture neural compression bias, while our proposed framework using classifiers, is able to highlight significant \textit{skin type} bias for images in the African racial group, supporting visual observation of image reconstructions.
    \item We reveal a phenotype-dependent correlation between bias and model architectures. Specifically, diffusion-based models exhibit severe \textit{skin type} bias for the African group, while the GAN-based model does not.
    \item Leveraging a racially balanced training dataset can reduce bias in certain cases but not in others, motivating further exploration into the development of balanced datasets and algorithmic bias mitigation methods.
    \item We can attribute two major sources of bias to (i) compression bias, in which a subject’s phenotype is genuinely altered (as confirmed by human annotations), stemming from phenotype imbalance in the dataset, and (ii) classification bias, where classifiers exhibit inconsistent error rates for different skin colors when images become blurred.
\end{itemize}

\section{Related Work}\label{sec:related}
\paragraph{Fairness in Image Compression}
Our work is closely related to~\citet{yucer2022does}, which studies the impact of JPEG compression on facial verification and identification tasks and the amount of adverse impact of JPEG compression on different racial and phenotype-based subgroups. They define bias as the different amount of downstream task performance degradation across groups. They find phenotype groups of darker skin tones, wide noses, curly hair, and monolid eye shapes suffer the most adverse impact in the facial recognition tasks. \citet{hofer2024taxonomy} study neural compression model reconstructions through visual inspection and gives a taxonomy of ``mis-compressions'', which they define as errors in semantic information after neural compression. Our work not only studies bias in neural compression through visual inspection but also aims to capture bias in a structured and scalable approach through a facial phenotype classifier. We see this as a first step towards systematically evaluating and mitigating bias in neural image compression models.

\paragraph{Fairness in Image Denoising and Upsampling}
Stemming from the ``White Obama" problem, fairness has been explored across image upsampling, denoising, and superresolution models. \citet{menon2020pulse}, the authors of the original model which suffers from the ``White Obama" problem, conduct an investigation concluding the bias is likely induced during the creation of the StyleGAN which they adopt for their task. \citet{jalal2021fairness} design novel definitions of fairness for image upsampling tasks and highlight fairness-accuracy tradeoffs for these types of models. \citet{Tanjim_2022_BMVC} examine the disappearance of minority attributes such as eye-glasses and baldness during image-to-image generation. They also propose a contrastive learning framework to improve upon bias in existing image-to-image translation models. \citet{laszkiewicz2024benchmarking} aim to study and benchmark the fairness in face image upsampling, demonstrating bias when imbalanced datasets are used while training these upsampling methods. 

\paragraph{Fairness in Face Analysis}
The processing of facial images is utilized across various domains, including face recognition, facial biometrics, and facial expression recognition. Fairness in such systems is crucial and has been studied in various aspects of the face and biometric analysis~\citep{drozdowski2020demographic, vangara2019characterizing, serna2019algorithmic}. \citet{buolamwini2018gender} evaluated commercial gender classification tools and identified that darker-skinned females suffer from significantly higher misclassification rates than lighter-skinned males. \citet{klare2012face} found that various face recognition systems exhibited the poorest performance on cohorts comprising females, Black individuals, and those aged 18-30. Motivated by the imbalanced distribution of datasets used for facial expression detection, \citet{xu2020investigating} investigate biases across gender, race, and age groups, and propose methods to mitigate these biases in such models.
\section{Problem Definition and Methods}\label{sec:method}
Overall, our goal is to develop a framework to evaluate and quantify bias in neural compression image reconstructions. In Section 3.1 we provide an overview of neural image compression. In Section 3.2 we define a general bias metric to evaluate bias in neural compression reconstructions. In Section 3.3 we highlight a specific instance of the bias metric, using a phenotype classifier to examine bias. 

\subsection{Neural Image Compression}

Neural compression models consist of an encoder $g_\text{enc}: \mathcal{X} \to \mathcal{Z}$ and a decoder $g_\text{dec} : \mathcal{Z} \to \mathcal{X}$, each built from learnable network layers. For each input image $x \in \mathcal{X}$, the encoder is used to obtain the latent space output $z$, which is then quantized to $\hat{z}$ and compressed losslessly to a bitstream. This bitstream is then decompressed to $\hat{z}$ and passed through the decoder to provide the decoded image $\hat{x}$. Overall, the goal for neural compression models is to minimize
\begin{equation} \label{eq:general}
     \mathcal{D}(x, \hat{x}) + \lambda\mathcal{R} (\hat{z})
\end{equation}
where $\mathcal{D}(x, \hat{x})$ is the distortion, $\mathcal{R}(\hat{z})$ is the compression bitrate, and $\lambda$ acts as the Lagrange multiplier that balances the rate-distortion trade-off. Distortion is typically measured using the mean squared error between the original image and the reconstruction while the bitrate is bounded using the entropy of the quantized latent $\hat{z}$.

\subsection{Evaluating Bias in Neural Compression}
We first define a general framework to analyze the bias in neural compression models. Let $\mathcal{D}=\left\{(x_i, y_i, a_i)\right\}_{i=1}^n$ be our dataset, where $x_i \in \mathcal{X}$ is our image, $y_i \in \mathcal{Y}$ is a label corresponding to a physical attribute of the image, and $a_i \in \mathcal{A}$ is a protected attribute. Our goal is to examine how the quality of reconstructions of $x_i$ differ across $\mathcal{A}$. Given a pretrained encoder and decoder, we can obtain the reconstructed dataset $ \widehat{\mathcal{D}}(g_\text{enc}, g_\text{dec})=\left\{(\hat{x}_i, y_i, a_i)\right\}_{i=1}^n$ where $\hat{x} = g_\text{dec} ( g_\text{enc} (x) ) $. Let $\mathcal{L}(\mathcal{D}, \widehat{\mathcal{D}} (g_{\text{enc}}, g_{\text{dec}}))$ be a loss metric to evaluate the quality of the reconstruction (e.g. distortion metrics, downstream task performance). Note that this original dataset needed for some loss functions (e.g. MSE) but we will omit it when it is not needed (e.g., downstream accuracy). Now, let
\begin{equation}
    \mathcal{L}(\mathcal{D}, \widehat{\mathcal{D}} (g_{\text{enc}}, g_{\text{dec}})\lvert a) =  \mathcal{L}(\mathcal{D}, \widehat{\mathcal{D}}_a (g_{\text{enc}}, g_{\text{dec}})),
\end{equation}
be conditional loss function where $\widehat{\mathcal{D}}_a(g_{\text{enc}}, g_{\text{dec}}) = \{(\hat{x}_i, a_i, y_i) \in \widehat{\mathcal{D}}(g_{\text{enc}}, g_{\text{dec}}) | a_i = a\}$.
Using this conditional loss, we can define bias as:
\begin{equation} \label{eq:general_bias}
    \text{Bias} \triangleq \max_{a,b \in A}[\mathcal{L}(\widehat{\mathcal{D}}(g_{\text{enc}}, g_{\text{dec}})|a) - \mathcal{L}(\widehat{\mathcal{D}}(g_{\text{enc}}, g_{\text{dec}})|b)].
\end{equation}
This bias term represents the maximum difference in loss across groups in $\mathcal{A}$. As we will show in the following sections, the choice of loss function is crucial to reveal racial biases---traditional distortion metrics show little to no bias, while the accuracy of a phenotype classifier highlights significant bias across racial groups (Section~\ref{sec:bias_in_distortion}).

\subsection{Bias Evaluation with a Phenotype Classifier}\label{ssec:problem_definition}
From visual inspection of image reconstructions, we identify key facial phenotypes (e.g., skin color, eye shape) can get degraded under low-rate neural compression. To systematically quantify phenotype degradation induced by the neural compression architecture, accurate labels are required for image reconstructions. Hand-labeling the phenotypes in the reconstructed images would be the most accurate way to obtain these labels, but it is not a scalable procedure for large image datasets. Therefore we propose to use a neural-network-based phenotype classifier as a proxy of human evaluation. Additionally, using a classifier to identify biases across different racial groups offers valuable insights into the potential disparities that may emerge when reconstructed images are used in subsequent deep-learning tasks. While previous studies~\citep{jalal2021fairness, Tanjim_2022_BMVC, laszkiewicz2024benchmarking} have investigated the use of phenotype classifiers to assess or mitigate bias in facial images, we acknowledge that classifiers can be a source of bias as well, and we will discuss further in Section~\ref{ssec:source_of_bias}.

First, given a dataset $\mathcal{D}$ where $\mathcal{A}$ is the set of racial groups (e.g $\{$African, Asian, Caucasian, Indian\}), and $\mathcal{Y}$ is the set of possible phenotype labels (e.g $\{$bald, curly hair, straight hair, wavy hair\} for \textit{hair type}), we split into $\mathcal{D_\text{train}}$ and $\mathcal{D_\text{test}}$ and use $\mathcal{D_\text{train}}$ to train a classifier $f: \mathcal{X} \to \mathcal{Y}$ to predict the phenotype labels (this can be a binary or multiclass classification task).  Then, given a pretrained encoder and decoder at bitrate $r$, the original test dataset $\mathcal{D_\text{test}}$ is compressed to the bitrate $r$ and reconstructed to $\mathcal{\widehat{D}}^r_\text{test}(g_\text{enc}, g_\text{dec})=\left\{(\hat{x}_i, y_i, a_i)\right\}_{i=1}^n$. 
To measure phenotype degradation at the given rate, we define our loss function to be the error rate of $f$ on $\mathcal{\widehat{D}}^r_\text{test}$: 
\begin{equation}
    \text{Err}(\mathcal{\widehat{D}}^r_\text{test}(g_\text{enc}, g_\text{dec})) = \mathbb{P}_{(\hat{x}, y) \sim \mathcal{\widehat{D}}^r_\text{test}(g_\text{enc}, g_\text{dec})}(f(\hat{x}) \neq y).
\end{equation}
The conditional loss then becomes:
\begin{equation}\label{eq:cond_acc}
    \text{Err}(\mathcal{\widehat{D}}^r_\text{test}(g_\text{enc}, g_\text{dec}) \lvert a)
    = \mathbb{P}_{(\hat{x}, y, a) \sim \mathcal{\widehat{D}}^r_\text{test}(g_\text{enc}, g_\text{dec})}(f(\hat{x}) \neq y \lvert A = a).
\end{equation}

By defining the loss function to be the error rate of the phenotype classifier, our bias metric directly becomes \textit{accuracy disparity}, the maximum difference of accuracy across all groups (due to the standard relationship between error rate and accuracy). Given a rate $r$, an encoder $g_\text{enc}$, and a decoder $g_\text{dec}$, the bias metric is defined as:
\begin{equation}\label{eq:bias}
    \text{Bias}(\mathcal{\widehat{D}}^r_\text{test}(g_\text{enc}, g_\text{dec}))
    \triangleq \max_{a, b \in \mathcal{A}} [\text{Acc}(\mathcal{\widehat{D}}^r_\text{test}(g_\text{enc}, g_\text{dec}) | a) - \text{Acc}(\mathcal{\widehat{D}}^r_\text{test}(g_\text{enc}, g_\text{dec}) | b)]
\end{equation}
 
where $\text{Acc}(\mathcal{\widehat{D}}^r_\text{test}(g_\text{enc}, g_\text{dec}) | a) = 1 - \text{Err}(\mathcal{\widehat{D}}^r_\text{test}(g_\text{enc}, g_\text{dec}) | a)$. This definition of bias is derived from a popular fairness metric, \textit{accuracy parity}, in which equal accuracies across all groups imply fairness in a classifier \citep{berk2017fairness, zafar2017fairness}. The motivation behind the selection of this bias definition can be observed in the following example.

\begin{example} \label{example}
    Let $\mathcal{A}$ be the set of races $\{$African, Caucasian$\}$ and let $\mathcal{Y} = \{$light skin, dark skin$\}$. In this case, the conditional error in Equation \ref{eq:cond_acc} captures the error rate of the skin color classification in the reconstructed image space for each group. When these conditional error rates are similar across $\mathcal{A}$, the skin colors switch equally for both groups in $\mathcal{A}$. When these values are different across $\mathcal{A}$, one race suffers from a skin color switch significantly more than another. Thus, the bias metric presented in Equation \ref{eq:bias} captures a more descriptive insight into what leads to race flipping than traditional metrics, which may only capture the frequency of the race flipping~\citep{jalal2021fairness}. By changing $\mathcal{Y} = \{$monolid eyes, non-monolid eyes$\}$ or any other phenotype, we can gain additional insight into how specific phenotypes get lost at different rates across each group in $\mathcal{A}$.
\end{example}
\section{Experiments and Evaluation}\label{sec:evaluation}
\subsection{Experimental Setup} \label{ssec:exp_setup}
\paragraph{Neural Compression Models}
In this paper, we evaluate a diverse collection of neural image compression models across different bitrates. An overview of our models is shown in Table \ref{model_table}. 
We evaluate three fixed-rate models, \textit{Hyperprior}~\citep{balle2018variational}, \textit{Joint}~\citep{minnen2018joint}, and \textit{GaussianMix-Attn}~\citep{cheng2020learned}. All of these models are trained towards a fixed trade-off between rate and distortion as highlighted in Equation \ref{eq:general}. We train these models to five operational bitrates. 
The model proposed in the \textit{QRes} paper~\citep{duan2023lossy} is a progressive decoding model that supports encoding images to 12 bitrates with one trained model. This is achieved by encoding only a subset from all the available latent variables. We follow this approach and encode images to 5 different bitrates with progressive decoding.
The \textit{VarQRes} model~\citep{duan2023qarv} is a variable rate compression model. The network is trained to operate in a range of rate-distortion trade-off points. Additionally, we consider two models which leverage attributes of popular generative models. The \textit{HiFiC} model~\citep{mentzer2020high} combines GANs with neural compression by introducing a discriminator conditioned on the latent variable following the decoder. The \textit{CDC} model~\citep{yang2023lossy} is a conditional diffusion model which closely resembles a diffusion-based autoencoder. In addition to the standard \textit{CDC} model, we consider two variants, \textit{CDC-L2} in which an auxiliary loss term is added that directly captures the distortion between the original image and the generated image, and \textit{CDC-LPIPS}, where the model adds an optional realism loss measured by LPIPS~\citep{zhang2018unreasonable}. We describe model implementations and training details in Appendix~\ref{appendix:nic}.

\paragraph{Phenotype Classifier} To study phenotype degradation in decoded images from neural compression, we use the Racial Faces in the Wild (RFW) dataset~\citep{wang2019racial} and a recently released facial phenotype annotation dataset specifically for RFW~\citep{yucer2022measuring}. This annotation dataset provides labels for six phenotype categories—skin type, eye type, hair type, hair color, lip type, and nose type—across four racial groups: African, Indian, Asian, and Caucasian. Skin types are labeled into 6 classes according to Fitzpatrick Skin Types~\citep{fitzpatrick1988validity}. Eye types are labeled as monolid or non-monolid. Nose types are labeled wide or narrow depending on nasal breadth. Hair types are labeled into 4 groups: bald, curly, straight, and wavy. Lip types are labeled as either full or small. Hair colors are labeled red, grey, black, blonde, and brown. The distribution of phenotypes across these racial groups is depicted in Figure~\ref{fig:phenotype-dist}.

We train individual classifiers for each phenotype classification task (e.g. one model for eye type classification, one for hair type classification, etc.), leading to either a binary or multi-class classification task. Training details for the phenotype classifiers can be found in Appendix~\ref{appendix:phenotype_classifier}. When measuring bias, we utilize the racial groups as our sensitive attribute, defining $\mathcal{A}$ as the set of all racial groups. When performing inference for multi-class classification tasks hair color and hair type, we group the three most dominant classes for each group. For skin type, we group all classes that make up at least 5\% of the group. This allows us to evaluate the extent to which phenotypes flipped to those not prevalent in the racial group of the raw image.

\paragraph{Datasets} We train all neural compression models on the CelebA~\citep{liu2018large}, FaceARG~\citep{darabant2021FaceARG}, and FairFace \citep{karkkainen2019fairface} datasets. These datasets are chosen to make comparisons between the impact of racially balanced and imbalanced training sets. The CelebA dataset has a significantly imbalanced racial composition with more than 70\% of the images from the white racial group \citep{karkkainen2019fairface}. Additionally, we leverage the FaceARG dataset and the FairFace Dataset to investigate the effect of a balanced training dataset. FaceARG is a large-scale dataset containing over 175,000 facial images, each labeled with age, gender, race, and ethnicity. The dataset features a relatively balanced distribution of images across four different racial groups: African, Asian, Caucasian, and Indian. The FairFace dataset contains over 100,000 images with a balanced racial composition across seven race groups: White, Black, Indian, East Asian, Southeast Asian, Middle Eastern, and Latino. All images are down-sampled to 64x64 resolution. Finally, to quantify the relationship between realism and bias, we utilize the DemogPairs dataset \citep{hupont2019demogpairs} as a reference to compute FID scores of the decoded images.

\begin{figure}[t]
    \centering  
    \includegraphics[width=0.8\columnwidth]{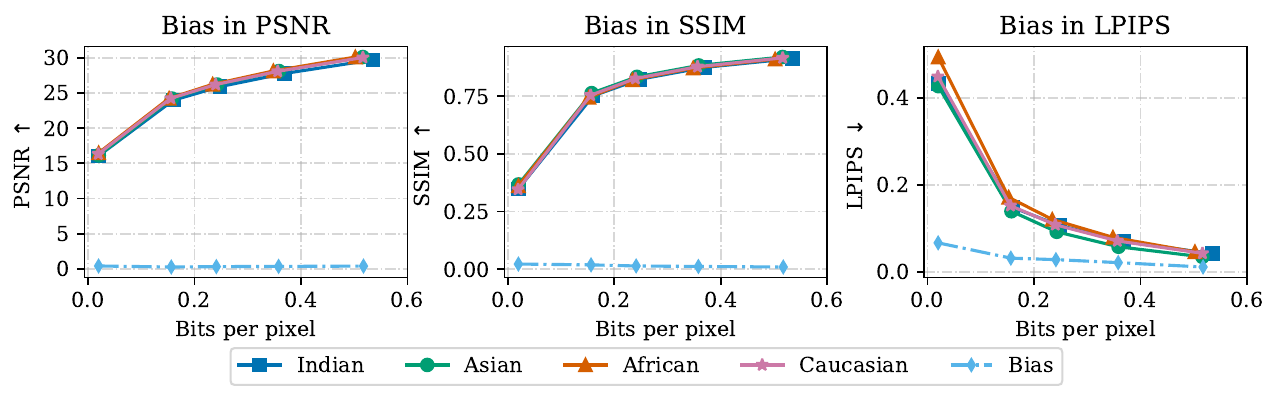}
    \caption{
    Traditional rate-distortion metrics (PSNR, SSIM, and LPIPS) for the \textit{Joint} model trained on the CelebA dataset, shown for each race and the overall dataset. The rate-distortion curves are nearly identical across all races for PSNR and SSIM, which contrasts with the findings from the qualitative analysis. While the LPIPS curve for the African group is slightly higher than for other races, it fails to fully reflect the disparities observed in the qualitative analysis.
    }
    \label{fig:per-race-rate-distortion}
\end{figure}

\subsection{Do neural compression models exhibit bias? How can we quantify it?}\label{sec:bias_in_distortion}
In Figure~\ref{fig:intro_visualization}, we observed examples of compressed images showing potential biases on non-Caucasian faces. We aim to systematically quantify such biases by employing different loss functions in our bias definition (Equation~\eqref{eq:bias}). We show that while all traditional distortion metrics fail to capture any biases, phenotype classification metrics uncover significant disparities related to skin type, eye type, and hair type.

\paragraph{Traditional Distortion Metrics} 
We first examine distortion metrics commonly used in compression: Peak Signal-to-Noise Ratio (PSNR), Learned Perceptual Image Patch Similarity (LPIPS) and Structural Similarity Index Measure (SSIM). We provide rate-distortion curves using these metrics using 
the \textit{Joint} model trained on the CelebA dataset in Figure~\ref{fig:per-race-rate-distortion}. Rate-distortion plots for other models are presented in Appendix~\ref{appendix:RD}. The rate-distortion curves highlight that distortion values across each race are nearly identical to that of the overall dataset, suggesting that facial images in different race groups are distorted by similar amounts at similar rates. The LPIPS curve for African faces sits slightly higher than the others but does not capture the extent of change seen in the qualitative analysis. This indicates that \emph{traditional distortion metrics are not suitable for capturing the bias in these neural compression architectures}, which motivates the need for an alternative metric to capture this bias more effectively. 

\paragraph{Phenotype Classification Metric} 
We show a similar rate-distortion curve, where the $y$-axis is phenotype classification accuracy instead of traditional distortion metrics such as PSNR. We again show the result of  the \textit{Joint} model trained on the CelebA dataset in Figure~\ref{fig:first} as a representative example as results on all other models follow the same trend (see   Appendix~\ref{appendix:racial_bias}). The figure reveals a significant decline in classification accuracy for individuals in the African group at low bitrates, while accuracy for images from other racial groups remains relatively stable. This disproportionate drop in accuracy for the African group leads to an increased level of bias as the bitrate decreases, aligning with our qualitative analysis. 

In Figure~\ref{fig:second}, we plot bias values at different bit rates of the \textit{Joint} model for all six phenotypes. We observe that the bias in the classification of \textit{skin type}, \textit{eye type}, and \textit{hair type} increases as compression rates decrease, while other phenotypes display relatively low bias throughout. 
Specifically, the rise in bias for \textit{skin type} and \textit{eye type} is primarily driven by a disproportionate drop in accuracy for the African group, while the increased bias for \textit{eye type} is linked to a decline in accuracy for the Asian group. This bias trend is consistently observed, to varying degrees, across all other neural compression architectures studied. \emph{These findings suggest that the bias in compression—a disproportionate loss of racial identity in certain racial groups—can be effectively captured using skin color, eye type, and hair type phenotype classifiers.}

\begin{figure}%
\centering
\subfigure[]{%
\label{fig:first}%
\includegraphics[width=0.4\columnwidth]{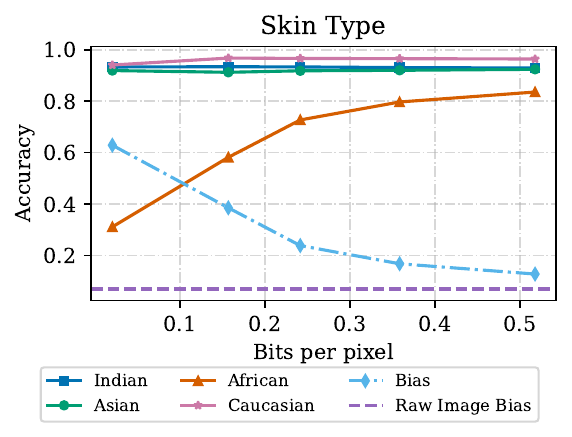}
}%
\qquad
\subfigure[]{%
\label{fig:second}%
\includegraphics[width=0.4\columnwidth]{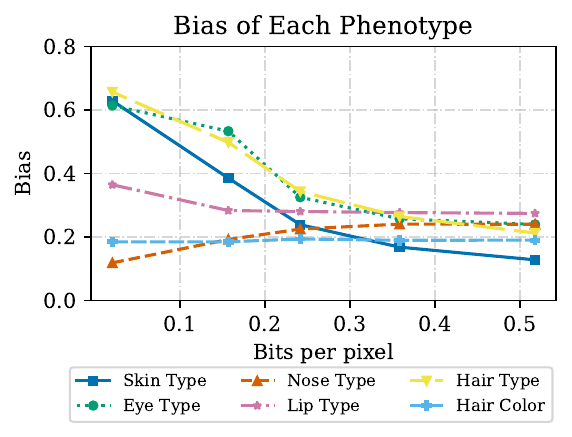}}%

\caption{(a) Bias for \textit{Skin Type} across different races for \textit{Joint}
reconstructions trained on the CelebA dataset. (b) As the bitrate is lowered, bias increases
for \textit{Skin Type}, \textit{Eye Type}, and \textit{Hair Type}, while remaining
relatively level for other phenotypes. }
\end{figure}

\subsection{How does bias vary across model architectures?}\label{ssec:bias_discussion}

\begin{figure}[t]
    \centering
    \includegraphics[width=0.9\columnwidth]{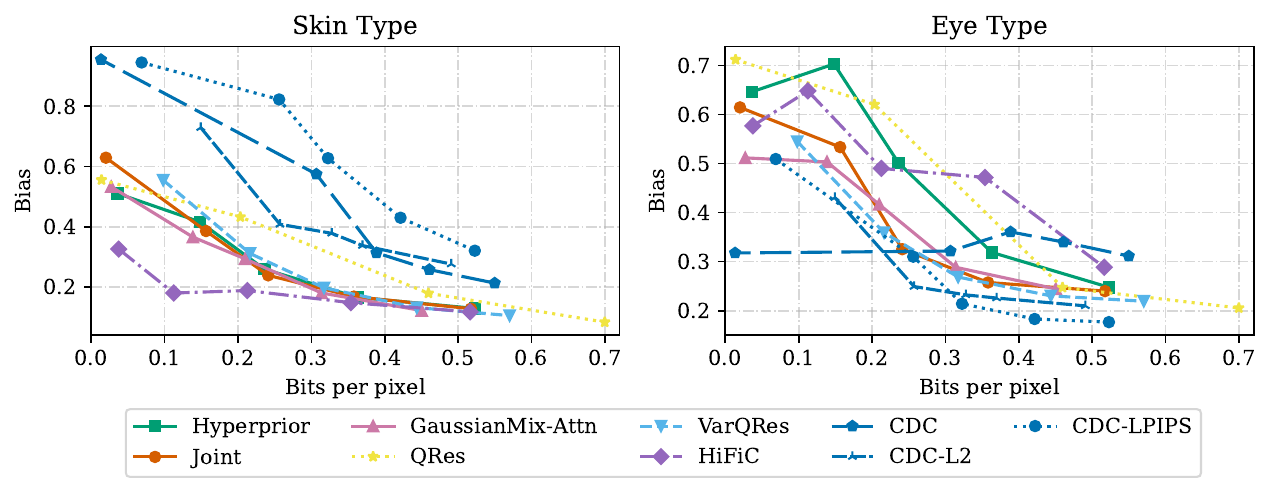}
    \caption{Bias in \textit{Skin Type} and \textit{Eye Type} across all neural compression models.}
    \label{fig:skin_type_bias_all_models}
\end{figure}

While we observed that across all models, significant biases appear in skin type, eye type, and hair type, we now want to delve deeper into how bias arises differently in the compression models we tested, especially between VAE, GAN, and diffusion-based models. To investigate this, we highlight the bias for different models for the \textit{skin type} and \textit{eye type} classification task in Figure~\ref{fig:skin_type_bias_all_models}. First, we observe that in the \textit{skin type} classification task, there is a clear relationship between the model architecture and the bias we observe. The diffusion models (\textit{CDC}, \textit{CDC-L2} and \textit{CDC-LPIPS}) appear to suffer from the most significant bias for the \textit{skin type} classification task, followed by the VAE-based models (\textit{Hyperprior}, \textit{Joint}, \textit{GaussianMix-Attn}, \textit{QRes}, and \textit{VarQRes}), and then the GAN-based model (\textit{HiFiC}). This data supports the visual observations we make from the reconstructed images from the \textit{HiFiC} model presented in Figure~\ref{appendix:hific}. Similar trends appear for the \textit{hair type} phenotype, with the exact rate-bias relationship plots in Appendix~\ref{appendix:hair_type_bias_across_models}.

This architecture dependence trend reverses when we explore \textit{eye type} classification. Here, the diffusion-based models experience the lowest amplification of bias while the GAN-based model experiences the highest level of bias. Again, the VAE-based models remain in the between the two types of generative models. These results suggest that the bias that different architectures vary across different classification tasks. We believe that future work can explore which specific properties of these model lead to specific types of bias and examine how leveraging properties from these architectures can help mitigate bias. 

\paragraph{Bias-Realism Relationship} In addition to directly comparing the bias, we systematically assess the relationship between bias and realism across neural compression models. This helps us understand whether models trade off these values and identify which objective each model can optimize. We quantify realism using Frechet Inception Distance (FID)~\citep{heusel2017gans}, while bias values are derived from Equation $\ref{eq:bias}$. FID provides statistical insight into how similar generated data is to a reference distribution. The reference distribution for FID is a set of real images to help capture how ``realistic" the decoded images are. To ensure we are measuring realism with respect to general facial datasets, we utilize the Demogpairs dataset \citep{hupont2019demogpairs} as a reference for computing FID. This enables us to capture the fidelity of the reconstructions without spurious correlations to any of the datasets used during training.

\begin{figure}[t]
    \centering
    \includegraphics[width=0.9\columnwidth]{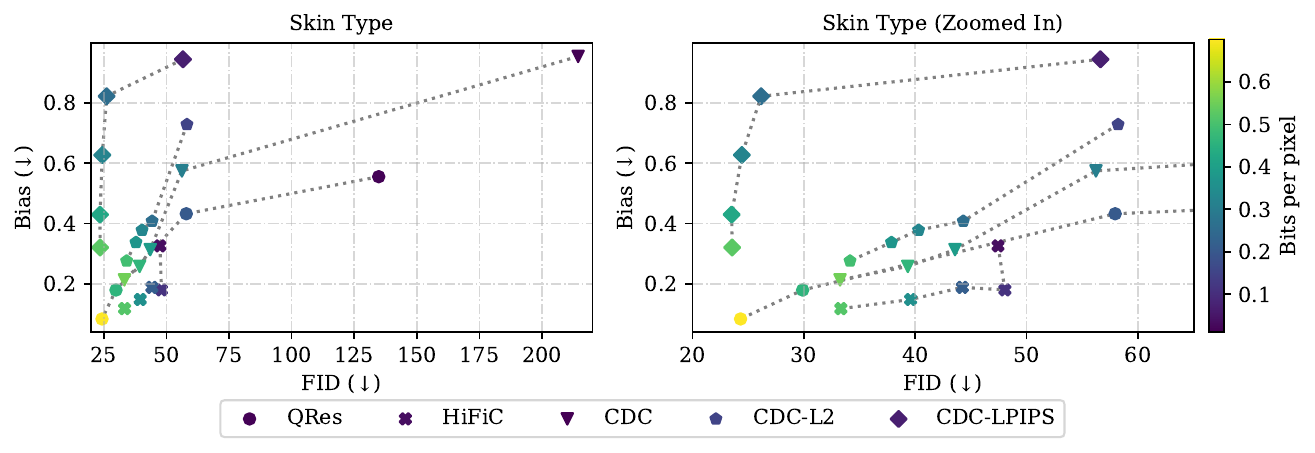}
    \caption{
    At high bitrates ($>0.1$ bpp), bias and realism are correlated across all the models. At low bitrates ($<0.1$ bpp), the trend is more sporadic.}
    \label{fig:fid-vs-bias}
\end{figure}

We highlight that at lower FID values, there appears to be a positive correlation between bias and realism (Figure~\ref{fig:fid-vs-bias}). As the realism deteriorates (FID increases), bias increases. These points mainly come from the intermediate bitrate regime. In the low bit rate regime, this trend degrades. Here, the relationship between bias and realism becomes much more sporadic. At lower levels of FID (higher realism), we can more clearly explore the relationship different neural compression models. We observe that \textit{CDC-LPIPS} is able to preserve realism well as the bitrate is reduced while its accuracy is significantly increased. The trend for the other models appear to be flatter and more linear indicating the positive correlation we observed in the original plot. We believe the bias-realism relationship suggests that future neural compression models should consider how to balance the increase of bias and loss of realism as compression bitrate decreases.

\subsection{Can using a balanced dataset remove the bias?} \label{sec:dataset_investigation}

\begin{figure}[t]%
\centering
\subfigure[]{%
\label{fig:celeba_vs_facearg}%
\includegraphics[width=0.44\columnwidth]{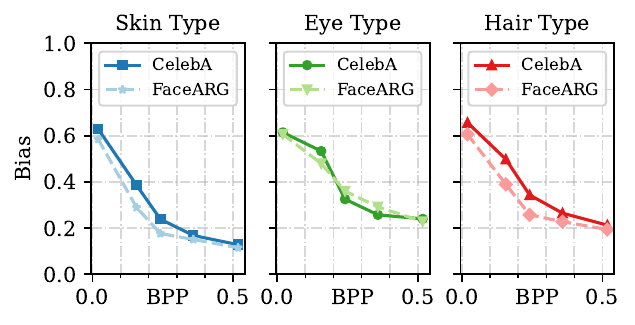}}%
\qquad
\subfigure[]{%
\label{fig:celeba_vs_fairface}%
\includegraphics[width=0.44\columnwidth]{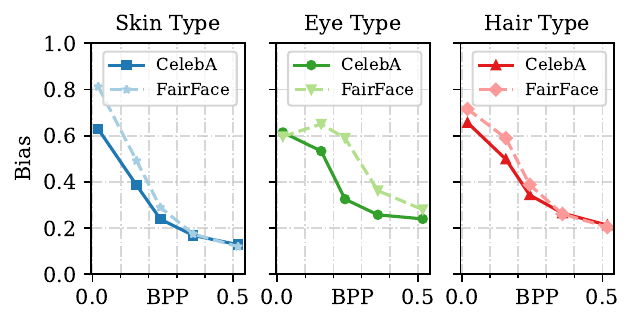}}%

\caption{(a) Using a racially balanced dataset (FaceARG) helps reduce the bias until extremely low bitrates less than 0.1 bpp. However, the general trend of bias increasing with decreasing bitrate is consistent across 2 datasets. (b) Using FairFace does not reduce bias and in cases increases bias. }
\end{figure}

As highlighted in Section \ref{ssec:exp_setup}, the CelebA dataset is infamously racially imbalanced, potentially leading to bias in downstream tasks. This motivates the exploration of utilizing a racially balanced dataset for training neural image compression models. We utilize the FaceARG dataset and the FairFace dataset to train our models and repeat our experiments from Section ~\ref{sec:bias_in_distortion}. First, we highlight scenarios in which training neural compression models with the FaceARG dataset reduces bias. As presented in Figure \ref{fig:celeba_vs_facearg}, the \emph{Joint} model trained on the racially balanced FaceARG dataset shows lower levels of bias in intermediate bitrates compared to the CelebA counterparts. This difference, however, is not explicit, and the trend of bias increasing with decreasing rates still exists. These trends slightly vary across other neural compression architectures and are presented in Appendix~\ref{appendix:balanced}. While bias is still present in this setting, these results suggest that leveraging a racially balanced training set for the neural compression model can reduce bias. 

However, leveraging another racially balanced dataset, FairFace, provides alternative insight. As we observe in Figure \ref{fig:celeba_vs_fairface}, the FairFace dataset does not improve, and in some cases increases bias, despite also being racially balanced. We highlight that this can be due to the imbalanced of the phenotype distribution within the races themselves. This lack of phenotype variability within racial groups can make certain phenotypes more difficult to preserve, which can lead to bias. This finding is consistent with that of~\citet{cherepanovadeepdive}, that class-balanced learning does not necessarily lead to fair classification. Additionally, the amplification of bias could be attributed to the facial orientation differences of the FairFace dataset \citep{laszkiewicz2024benchmarking}, in which images with more variable poses make reconstructions at lower rates, lower quality. We conclude that training with a balanced dataset can reduce bias in some cases but is not a sufficient bias mitigation strategy. We believe that this strongly motivates the construction of datasets that are balanced beyond race (e.g. phenotype level bias) to further reduce bias. Additionally, this motivates algorithmic methods for bias mitigation in neural image compression architectures, some of which we discuss in Section~\ref{sec:conclusion}.

\subsection{Why is my neural compression model discriminatory?}\label{ssec:source_of_bias}
\begin{figure}[t]
    \centering
    \includegraphics[width=0.5\columnwidth]{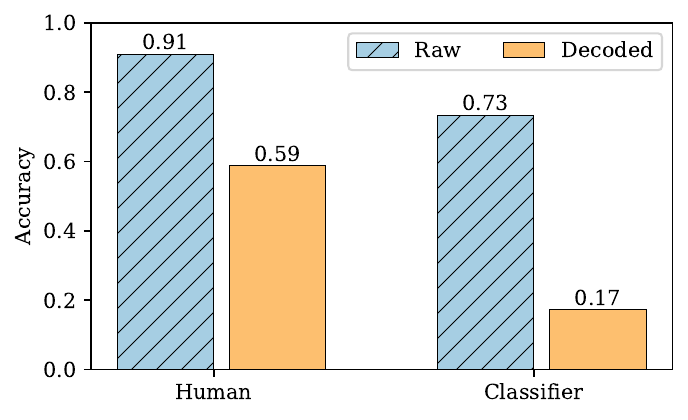}
    \caption{Both human and classifier models display reduced accuracy in labeling African skin types at low bit rates. }
    \label{fig:human_annotation}
\end{figure}

In Figure~\ref{fig:skin_type_bias_all_models} we observe that all neural compression models have substantial bias when the bitrate becomes lower. In other words, when the compression becomes very lossy, neural networks decide to lose more information about some racial groups than others. 
Why does this problem still persists when train neural compression models with a racially balanced dataset? We aim to explain the source of this discrimination by decomposing this into \textbf{compression model bias} and \textbf{classification model bias}. The compression model bias can come from dataset composition and model architecture. Since all models suffer from bias towards African racial group, we focus on the common training dataset. We aim to identify the impact of training dataset, and understand whether designing specific training set for African racial group helps. 
We start by looking into the compression model bias. Ensuring the accuracy of classifier is crucial in this process. To validate our findings, we acquired human annotations on a subset of images (details in Appendix~\ref{appendix:annotation}) to analyze how humans classify skin type in facial images. We asked annotators to label both original images and distorted images decoded from the lowest bitrate at around 0.02 bpp from the \textit{GaussianMix-Attn} model. From the human annotations, we observe human’s ability to discern the original skin type for African racial group drops 32\%. This reflects that humans perceive decoded African facial images as lighter, which is also consistent with the classifier accuracy results. As shown in Figure~\ref{fig:human_annotation}, the accuracy in human labellings drops less than the accuracy in classifier, suggesting there might be other factors that impact the classifier, which we discuss in more detail later in this subsection. 

\begin{figure}[h]%
\centering
\subfigure[]{%
\label{fig:african_only_2_models_visual}%
\includegraphics[width=0.55\columnwidth]{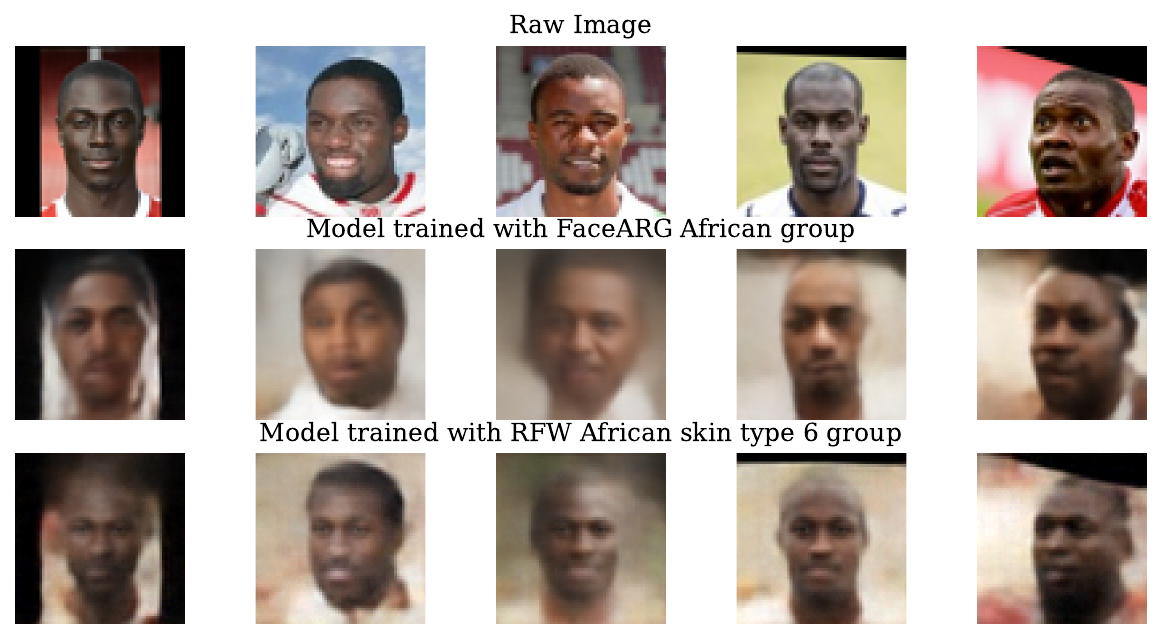}}%
\qquad
\subfigure[]{%
\label{fig:african_only_2_models_plots}%
\includegraphics[width=0.4\columnwidth]{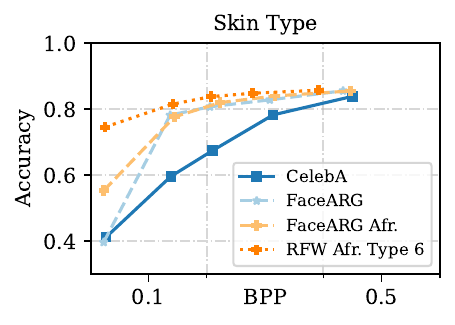}}%
\caption{(a) Compression model \emph{GaussianMix-Attn} trained with strictly African images of skin type 6 shows better capability at reconstructing the correct skin color, than the model trained on diverse African faces. (b) The bias for African skin type is contained at a low level by training with skin type 6 images. }
\label{fig:african_only_comparisons}%
\end{figure}

\textbf{Does training with exclusively African facial images help?}
Since training with a racially balanced dataset doesn't necessarily remove bias towards the African group, training exclusively on African group may provide insights into whether the source of bias is solely the dataset. Therefore, we exclude other racial groups in the FaceARG training set and evaluate the bias towards African group in this case. As shown by Figure~\ref{fig:african_only_2_models_plots}, even when training exclusively with African facial images, the bias still persists. The reason for this may be that the skin color composition within the African racial group is unbalanced. Although there are no direct skin type labels for the FaceARG dataset, the RFW dataset, as a reference, suggests there is 58\% of type 5 and 39\% of type 6 within the African racial group. The compression model is trained to fit to the majority within the group. This causes the decoded images of some African facial images to appear lighter than the original. 

\textbf{Can we further minimize the impact of training set skin type distribution?} With this goal in mind, we trained the \emph{GaussianMix-Attn} model with exclusively skin type 6 African facial images from the RFW dataset. In this case, we observe the decoded images align more with the original image, as illustrated by Figure~\ref{fig:african_only_2_models_visual}. The visual improvement is also supported by the classifier accuracy improvement, as shown in Figure~\ref{fig:african_only_2_models_plots}. By restricting the training set to first African only, then African with skin type 6 only, we can improve the skin type accuracy at the lowest rate, indicating the compression model bias has been reduced. 

Even though the classifier accuracy improves, it still classifies 30\% of images from skin type 6 to skin type 3. In both this case and the previous human annotations, the classifier experiences a greater accuracy drop than humans at the lowest rate. Both facts prompt us to explore additional contributions to the bias. Since human labeling is impacted only by the compression model bias, the difference in accuracy reduction would come from the additional bias introduced by the classifier. 

\begin{figure}[t]
    \centering
    \includegraphics[width=\columnwidth]{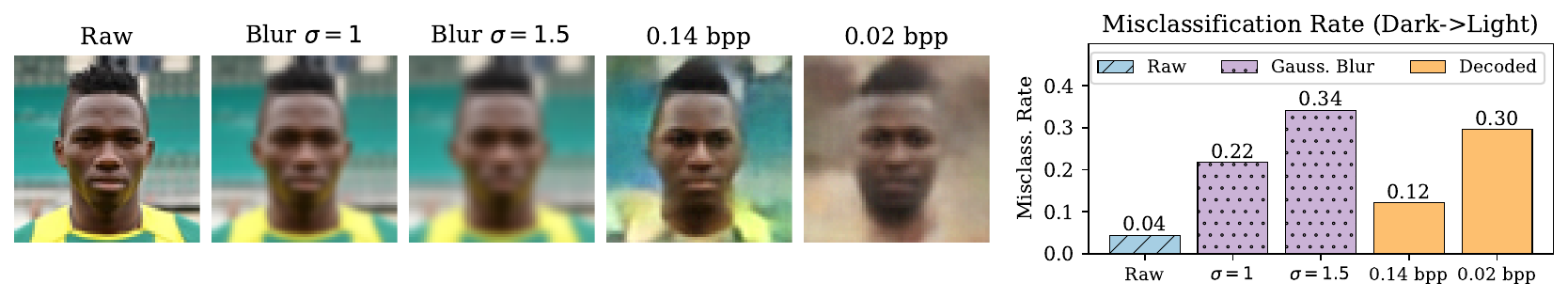}
    \caption{Gaussian blurring negatively impacts classifier accuracy, making classifier to mis-classify the darkest skin type 6 to lighter skin type 3. The percentage of images that are classified to type 3 from type 6 increases with the level of blurring. }
    \label{fig:gaussian_blur}
\end{figure}

\textbf{Is classifier introducing bias by looking at irrelevant image features?}
For VAE-based models, a dominant artifact they share at low rates is blurriness. It is therefore crucial to understand if blurriness alone has an impact on the skin type classifier. We evaluate the model's sensitivity to blurriness in decoded images. As shown in Figure~\ref{fig:gaussian_blur}, applying Gaussian blurring to the raw images will result in skin type classifiers mis-classify to skin type 3 for African racial groups. The percentage of skin type flipping to minority labels is no more than 6\%, while it is more than 30\% for African racial group. It is possible that the classifier is impacted by blurry inputs during classification while classifying for the skin type. This can explain the difference in accuracy reduction between the human labeling and the model shown in Figure~\ref{fig:human_annotation}.

\section{Conclusion and Discussion}\label{sec:conclusion}

We benchmarked nine state-of-the-art neural compression models—including VAE-, GAN-, and diffusion-based architectures—by assessing their bias when applied to low-rate facial image compression. Our analysis reveals substantial bias in phenotype preservation, especially for African individuals’ \textit{skin} and \textit{hair types} and Asian individuals’ \textit{eye types}. While all models exhibit a similar bias trend, they strike different trade-offs between realism and fairness (e.g., diffusion-based models offer high realism but amplify bias for African skin types). Furthermore, we find that simply balancing the racial composition of the training set does not suffice to eliminate this bias. Finally, through human annotations and additional experiments restricting the training data to specific phenotypes, we show that the root causes of bias stem from \textit{classification bias due to blurred images} and \textit{compression bias due to phenotype imbalance}. This work thus constitutes the first comprehensive study on bias in neural compression models.
\subsection{Broader Impact and Limitations}
\textbf{Broader Impact}
While our work centers on neural compression, the implications of this work extend beyond  compression applications. Low-rate neural compression models represent a direct implementation of a neural network with a highly constrained information bottleneck. The concept of an information bottleneck has been a compelling perspective for understanding neural network behaviors in areas such as classification~\citep{tishby2015deep}, representation learning~\citep{federici2020learning}, GANs~\citep{jeon2021ib}, and bandit learning~\citep{arumugam2021deciding}. Classification models, including facial recognition and anomaly detection, inherently exhibit an information bottleneck. Thus, the insights gained from studying neural compression models naturally extend to a wide range of neural networks that process human face images.

\textbf{Limitations} In this paper, we trained compression models with racially balanced dataset FairFace and FaceARG. However, it is hard to conduct this experiment with CelebA dataset due to lack of race labels. The racial composition of CelebA was found to be predominantly Caucasian~\citep{karkkainen2019fairface}. However, this finding was based on additional labelings on a subset of 3,000 images from 200k images. The absence of large-scale facial datasets with high-quality labels for both race and facial phenotypes prevented us from conducting highly controlled experiments. We advocate the community to push towards developing large scale datasets with high-quality labels for both race and facial phenotypes. 

In our study, the skin types for the RFW dataset were labeled following the Fitzpatrick skin type (FST), which has six types. Researchers have found FST to be racially limited~\citep{ware2020racial, heldreth2024skin}. Recently, there are new skin type measures such as the Monk Skin Tone (MST) scale~\citep{monk2019monk} that are designed to represent a broader range of people. Recent studies find that different skin type scales have impacts on inter-annotator agreements, as well as on an individual annotator's certainty~\citep{barrett2023skin}. Future work should consider looking into evaluating the bias in neural compression with various skin type measures, procedures, and datasets.

\subsection{Future Directions}
Although this paper focuses on identifying existing discriminatory harms in current neural compression models rather than mitigating bias, our work lays important groundwork toward developing mitigation algorithms. A key first step in any fairness mitigation approach is selecting the appropriate fairness metric to optimize. Our analysis demonstrates that straightforward extensions of accuracy parity—such as parity in mean squared error (MSE)—are inadequate for capturing racial bias in neural compression. Instead, we propose a classification-based metric that more effectively reflects such disparities. Besides, the analysis of the sources of bias provides a lens for potential mitigation strategies. A significant improvement in accuracy when using only one skin color phenotype suggests that decoupled, personalized models~\citep{dwork2018decoupled,ustun2019fairness} would benefit the minority group greatly in the neural compression application. However, this method faces some challenges. The biggest challenge is that most human face images do not come with detailed phenotype labels, and thus it is impossible to accurately divide the dataset to train each personalized compression model. Furthermore, as we divide the dataset to different phenotype intersections (e.g., dark skin, curly hair with a big nose), each group's sample size becomes smaller and the benefit of personalization diminishes~\citep{ustun2019fairness,monteiro2022epistemic} while the bit requirement to encode the group information grows. Thus, developing separate models for each phenotype group is only a partial, incomplete solution to the problem and underscores the need for an algorithmic approach to mitigate bias. We outline some potential future directions below. 

\textbf{Developing fair neural compression algorithms} 
First, since neural compression can be viewed as image-to-image models with information bottlenecks, an interesting future direction is exploring how traditional fair models from the standard image-to-image space \citep{Tanjim_2022_BMVC} translate to the neural compression domain. Another possibility could be to adopt bias mitigation techniques designed from representation learning \citep{zemel2013learning, louizos2015variational, creager2019flexibly} to the neural compression domain, as neural compression can be viewed as a rate-constrained version of representation learning. 
Other methods could explore leveraging components from fairness-aware generative models \citep{xu2018fairganfairnessawaregenerativeadversarial, friedrich2023fair} to design fair neural image compression models. Additionally, \citet{tschannen2018deep} proposes a distribution-preserving neural compression model, which, when combined with a racially balanced training set, could yield interesting insights into constructing a fair neural compression system.

\textbf{Isolating bias} For evaluation, we utilize a single phenotype classifier across different bitrates. This allows us to isolate the bias of the classifier by examining the performance differences across difference rates. Future work can further investigate isolating the bias of the phenotype classifier by leveraging a fair classifier. \citet{dooleyrethinking} demonstrate that bias can be inherent to the classifier architecture and that fair architectures can be found through neural architecture search. Exploring a fair architecture for neural compression is an interesting future direction. 
Additionally, emerging information theoretic techniques~\citep{goldfeld2021sliced,goldfeld2022k,wongso2022understanding,wongso2023using,tax2017partial,wibral2017quantifying,dutta2020information,dutta2023review} can be explored to further decouple bias in the encoder and decoder of neural compression architectures.

\bibliographystyle{ACM-Reference-Format}
\bibliography{references}

\appendix
\newpage
\counterwithin{figure}{section}
\counterwithin{table}{section}
\section{Dataset Details}
\begin{figure}[ht]
    \centering  
    \includegraphics[width=0.8\columnwidth]{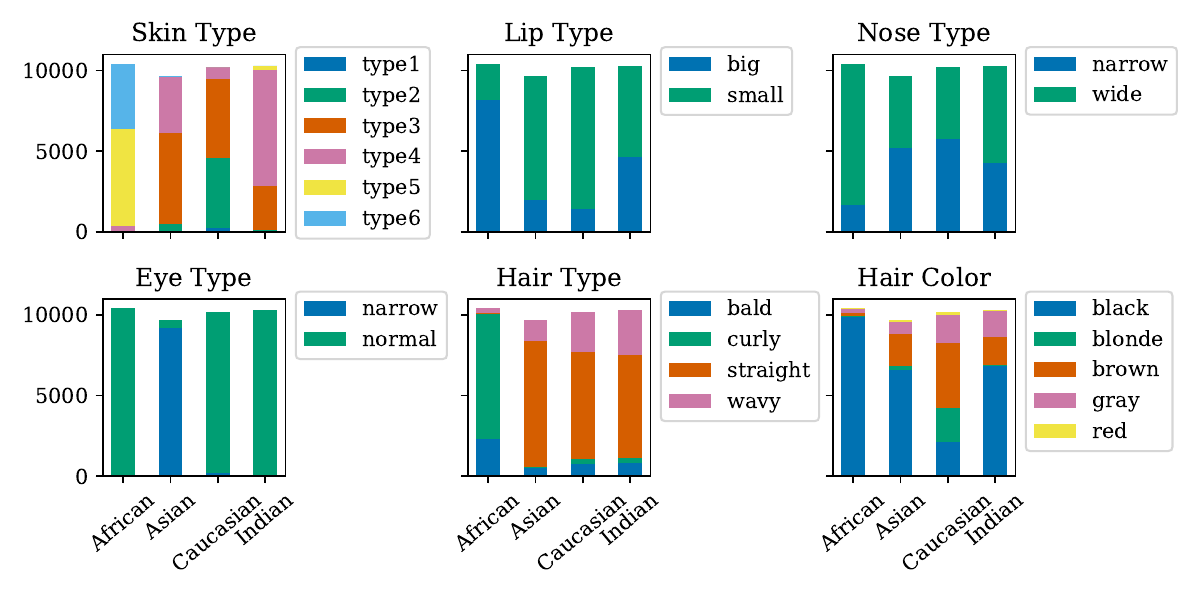}
    \caption{
    Distribution of phenotype classes for each category across racial groups in RFW dataset.
    }
    \label{fig:phenotype-dist}
\end{figure}

 We observe that across the dataset, phenotype distribution is imbalanced within each race. The distributions of \textit{skin type}, \textit{hair type}, and \textit{hair color} phenotypes are dependenet on racial group. The African group has predominantly type 5 and type 6 skin, curly hair, and black hair. The Asian group has predominantly type 3 and type 4 skin, straight hair, and black hair. The Caucasian group has predominantly type 2 and type 3 skin, with straight hair and a balanced hair color distribution. The Indian group has predominantly type 3 and type 4 skin, straight hair, and black hair. Additionally, the eye type labels are extremely imbalanced within each racial group with nearly all Asian images labeled as narrow and nearly all non-Asian images labeled as wide. The \textit{lip type} and \textit{nose type} distributions appear relatively balanced within each racial group.

\paragraph{Bias in Facial Image Datasets}
Machine learning models trained on biased datasets tend to inherit and perpetuate those biases, resulting in skewed performance across different demographic groups. Many large-scale facial image databases are disproportionately biased toward individuals with lighter skin tones, underrepresenting those with darker skin \citep{merler2019diversity}. For instance, widely used datasets like CelebA~\citep{liu2018large}, LFW~\citep{huang2008labeled}, and UTKFace~\citep{zhifei2017cvpr} reflect significant demographic imbalances. Beyond skin tone, other attributes such as gender and age are also prone to bias in representation. Numerous studies have explored how these biases in datasets affect the performance of downstream models, particularly in terms of fairness across demographic groups~\citep{drozdowski2020demographic, buolamwini2018gender, hupont2019demogpairs}. In response, recent efforts have focused on creating more diverse and discrimination-aware facial image datasets, such as FairFace~\citep{karkkainen2019fairface}, Racial Faces in-the-Wild (RFW)~\citep{wang2019racial}, and FaceARG~\citep{darabant2021FaceARG}, to reduce model biases and improve fairness. While these datasets reduce bias in terms of racial representation, they do not fully eliminate all forms of bias. In this paper, we focus on the facial phenotypes within the RFW dataset, which offers a relatively balanced racial composition. However, it remains imbalanced at the phenotype level, a limitation that will be explored in detail in the paper.

\section{Traditional Distortion Metrics}\label{appendix:RD}
\begin{figure}[h]
    \includegraphics[width=\columnwidth]{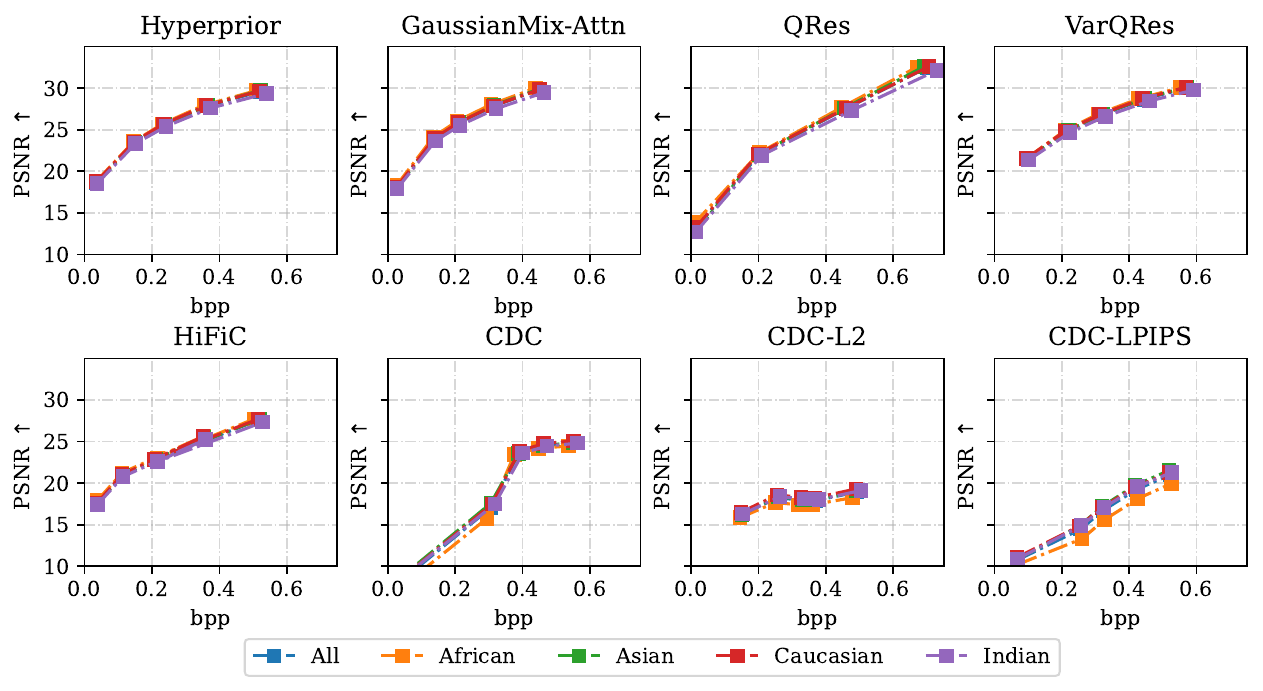}
    \caption{
    PSNR rate-distortion curves for all neural compression models trained on the CelebA dataset. 
    }\label{fig:psnr-all}
\end{figure}

\begin{figure}[h]
    \includegraphics[width=\columnwidth]{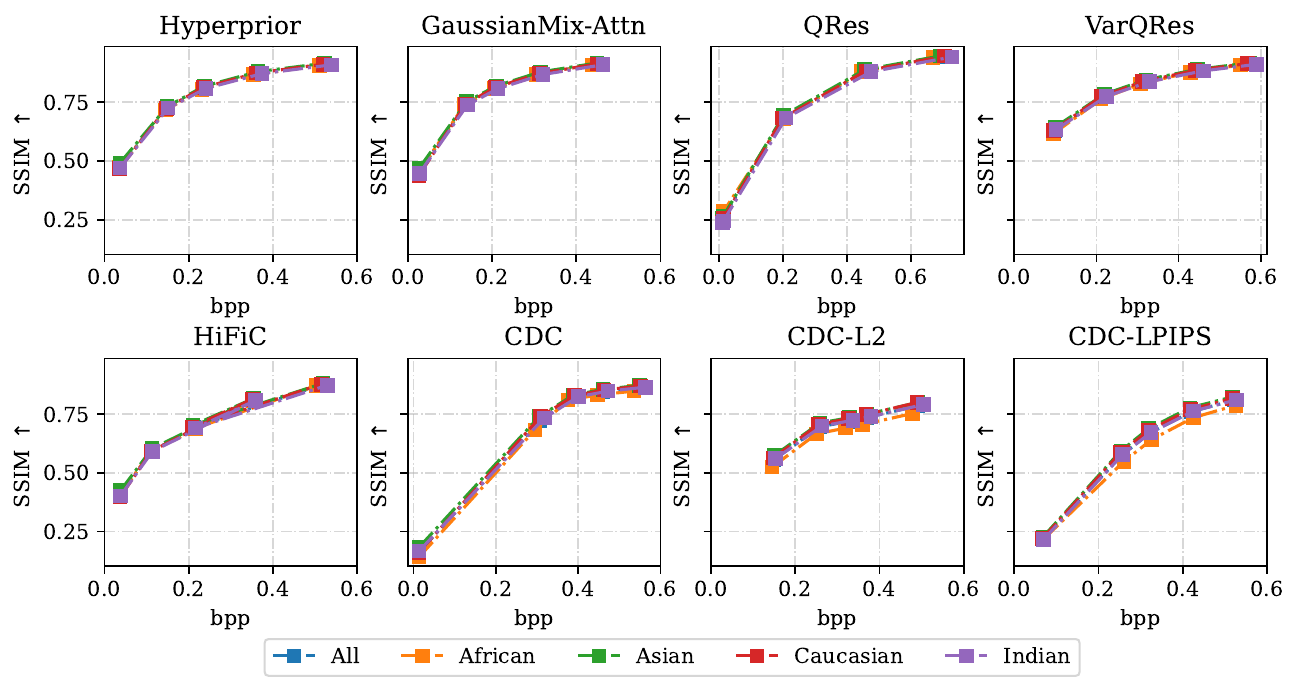}
    \caption{
    SSIM rate-distortion curves for all neural compression models trained on the CelebA dataset. 
    }\label{fig:ssim-all}
\end{figure}

\begin{figure}[h]
    \includegraphics[width=\columnwidth]{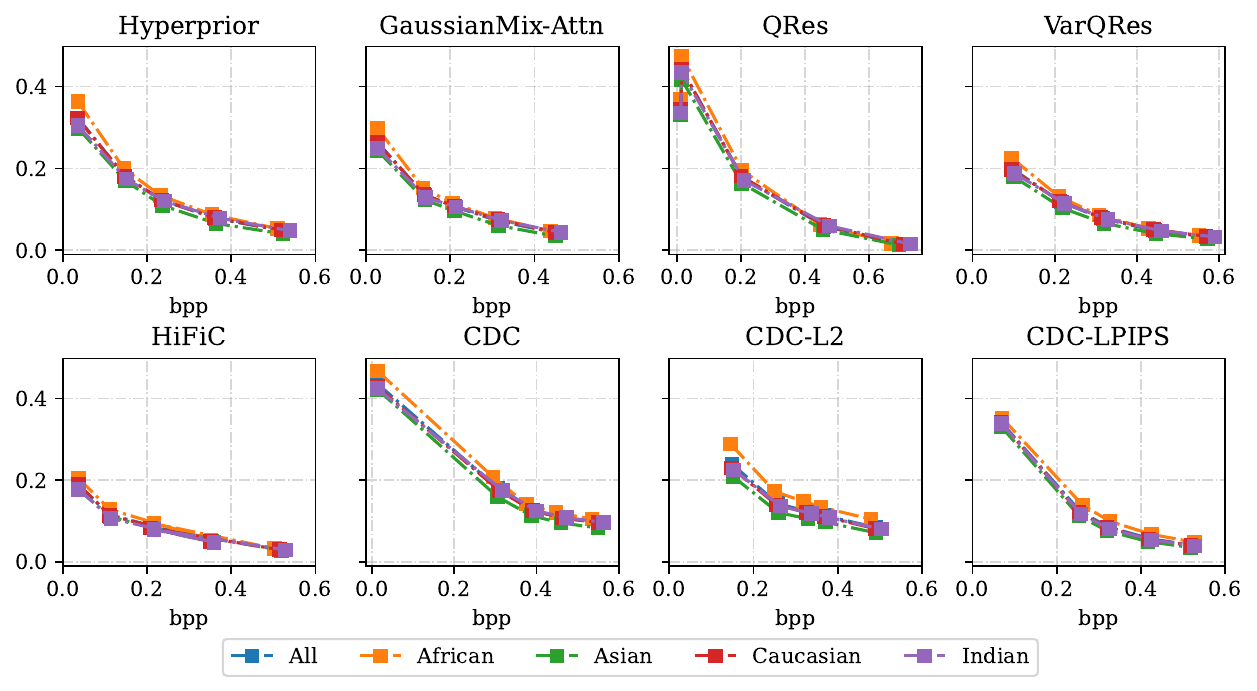}
    \caption{
    LPIPS rate-distortion curves for all neural compression models trained on the CelebA dataset. }
    \label{fig:lpips-all}
\end{figure}

We present the PSNR, SSIM, and LPIPS distortion curves for all models trained on the CelebA dataset in Figures \ref{fig:psnr-all}, \ref{fig:ssim-all}, and \ref{fig:lpips-all} respectively.

\clearpage
\newpage
\section{Training details}
\subsection{Phenotype Classifier}\label{appendix:phenotype_classifier}

We train ResNet18 models \cite{he2016deep} for facial phenotype classification from scratch. The classifiers retain the ResNet18 backbone and include a classification head for classifying the specific attribute. We trained the separate phenotype classifier models for up to 50 epochs, employing early stopping with patience of 5 epochs. We use cross entropy loss and optimize the models with the stochastic gradient descent optimizer, a fixed learning rate of 0.01, and a fixed batch size of 32. To evaluate each compression model at different compression rates, we train the models on decompressed images from each of the evaluated neural compression models with different compression rates separately, using the provided dataset annotations. We report the average results over 5 runs with different random seeds for all of our experiments.
\vspace{-4mm}
\subsection{Neural Compression Models} \label{appendix:nic}

For models \textit{Hyperprior}, \textit{Joint}, and \textit{GaussianMix-Attn}, we adopt the implementations from the CompressAI~\citep{begaint2020compressai} library.
For the other models, we adopt the implementation provided by the authors~\citep{duan2023lossy, mentzer2020high, yang2023lossy} or publicly available implementations~\footnote{https://github.com/Justin-Tan/high-fidelity-generative-compression}.
For the CompressAI neural compression models, we train for 1000 epochs with an early stopping patience of 50 epochs. We use a batch size of 64 and an initial learning rate of 0.0001. For the rest of the parameters, we leave them as they are implemented in the CompressAI repository. For the \textit{QRes}~\citep{duan2023lossy}, \textit{VarQres}~\citep{duan2023qarv}, \textit{HiFiC}\citep{mentzer2020high} and \textit{CDC}\citep{yang2023lossy} implementations, we follow the training procedure from the papers.
\clearpage
\newpage
\section{Racial Bias in Degradation} \label{appendix:racial_bias}
\begin{figure}[h!tbp]
    \includegraphics[width=\columnwidth]{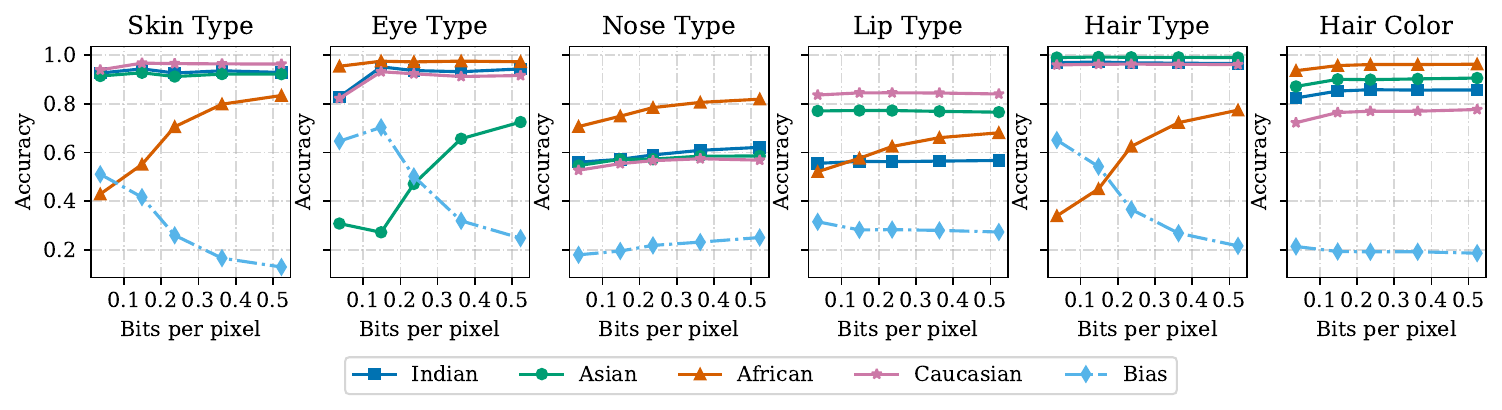}
    \caption{
    Bias in phenotype degradation for the \textit{Hyperprior} Model trained on CelebA
    }
\end{figure}

\begin{figure}[h!tbp]
    \includegraphics[width=\columnwidth]{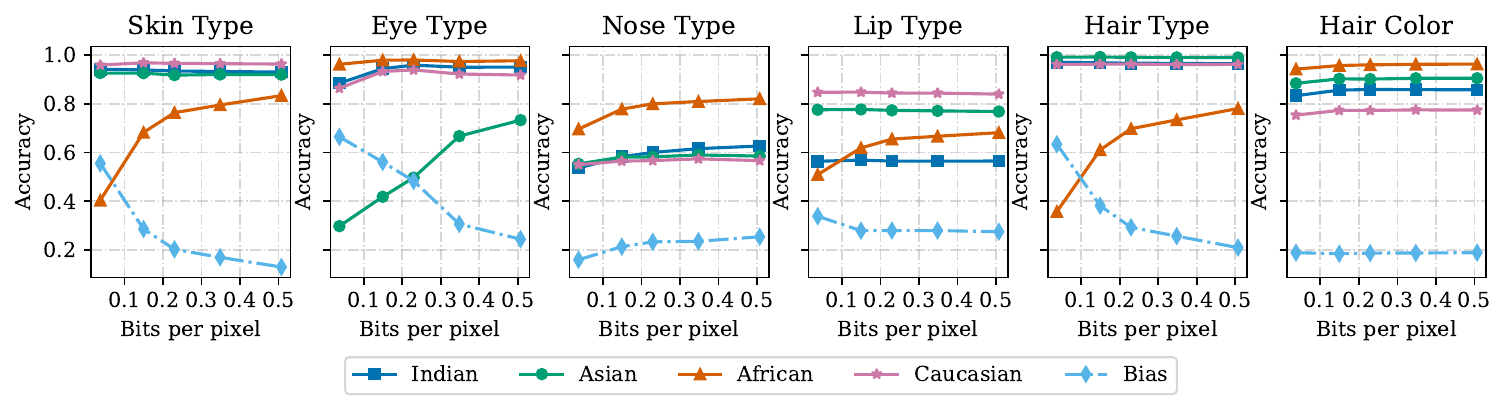}
    \caption{
    Bias in phenotype degradation for the \textit{Hyperprior} Model trained on FaceARG
    }
\end{figure}

\begin{figure}[h!tbp]
    \includegraphics[width=\columnwidth]{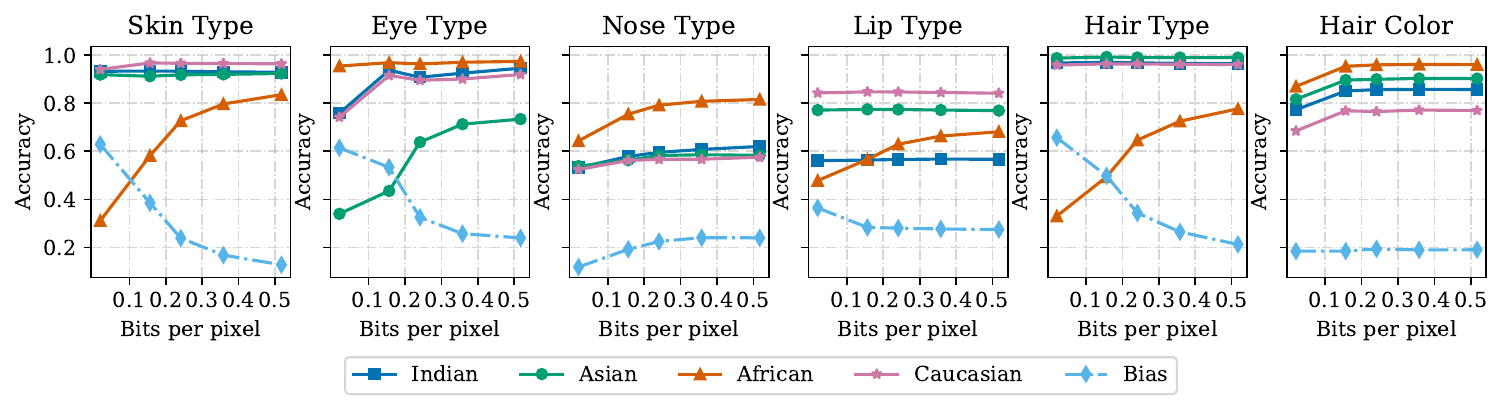}
    \caption{
    Bias in phenotype degradation for the \textit{Joint} Model trained on CelebA
    }

\end{figure}

\begin{figure}[h!tbp]
    \includegraphics[width=\columnwidth]{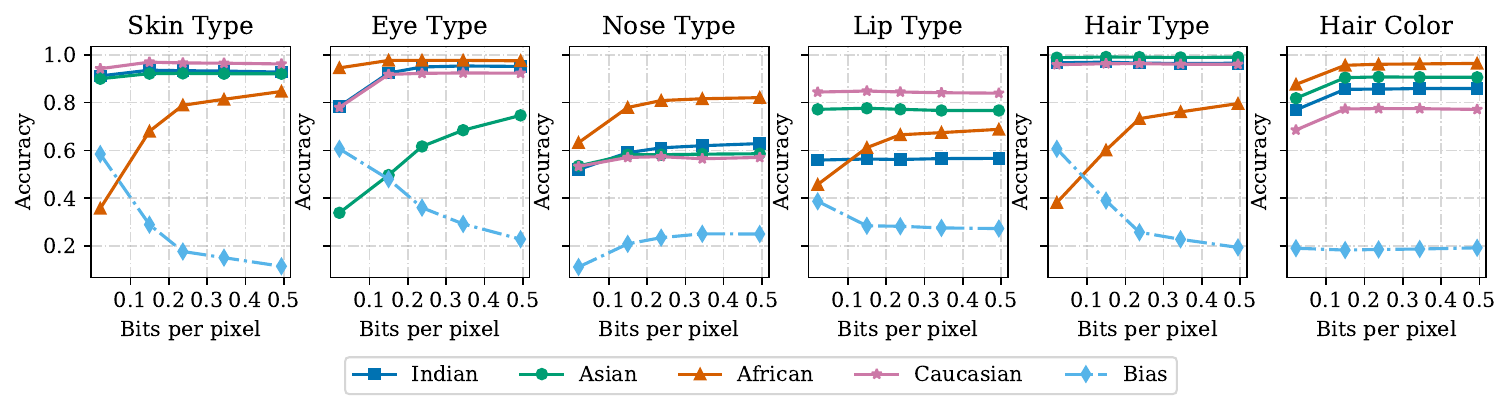}
    \caption{
    Bias in phenotype degradation for the \textit{Joint} Model trained on FaceARG
    }
\end{figure}

\begin{figure}[h!tbp]
    \includegraphics[width=\columnwidth]{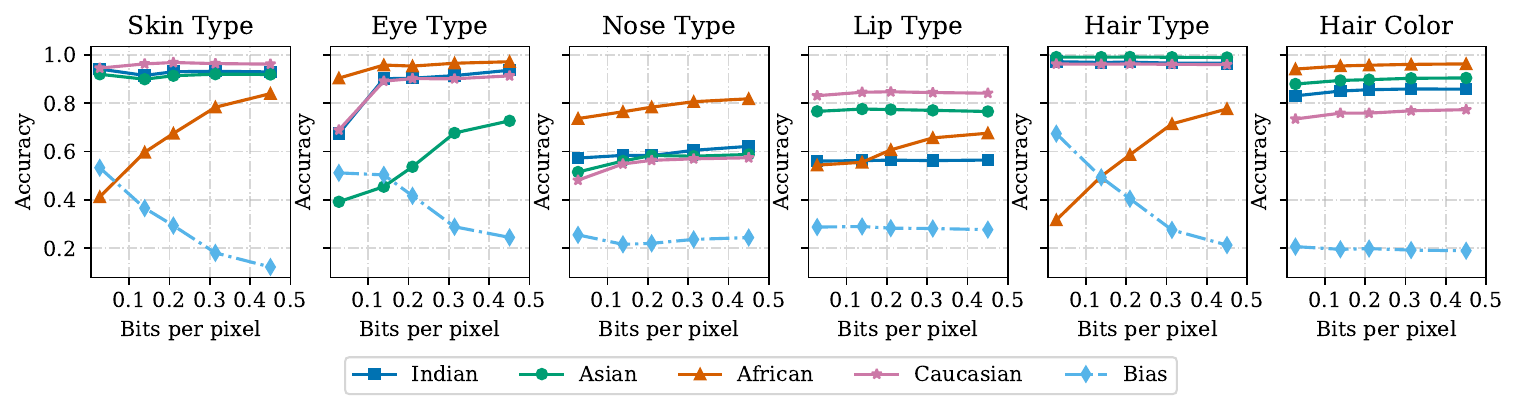}
    \caption{
    Bias in phenotype degradation for the \textit{GaussianMix-Attn} Model trained on CelebA
    }
\end{figure}

\begin{figure}[h!tbp]
    \includegraphics[width=\columnwidth]{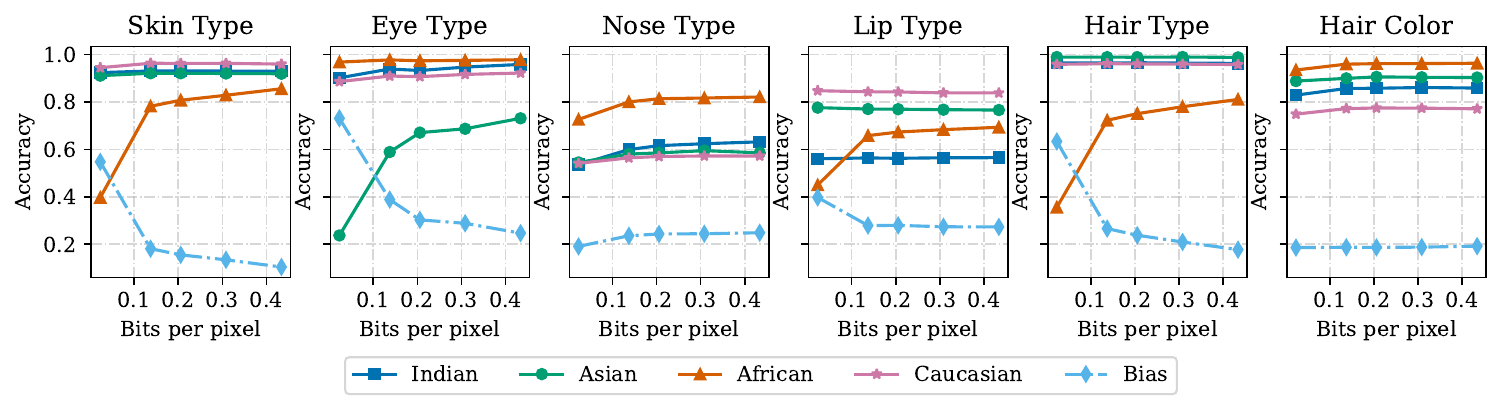}
    \caption{
    Bias in phenotype degradation for the \textit{GaussianMix-Attn} Model trained on FaceARG
    }
\end{figure}

\begin{figure}[h!tbp]
    \includegraphics[width=\columnwidth]{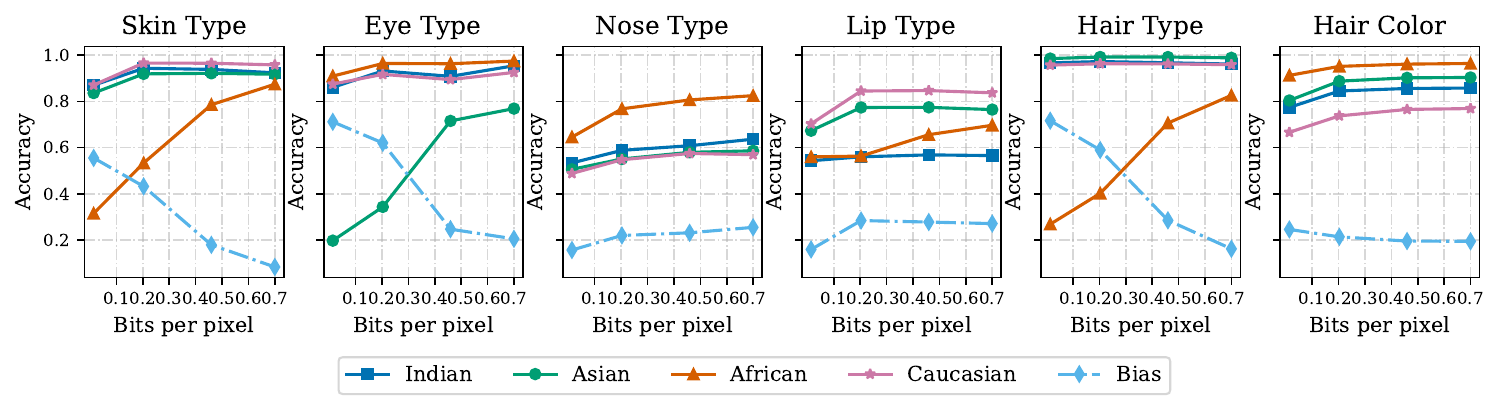}
    \caption{
    Bias in phenotype degradation for the \textit{QRes} Model trained on CelebA
    }
\end{figure}

\begin{figure}[h!tbp]
    \includegraphics[width=\columnwidth]{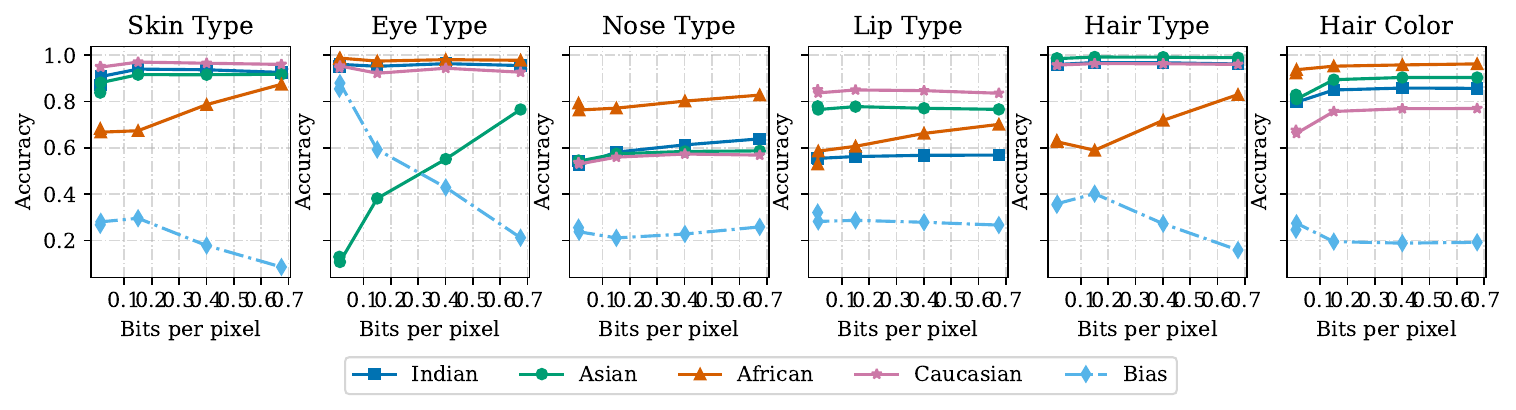}
    \caption{
    Bias in phenotype degradation for the \textit{QRes} Model trained on FaceARG
    }
\end{figure}

\begin{figure}[h!tbp]
    \includegraphics[width=\columnwidth]{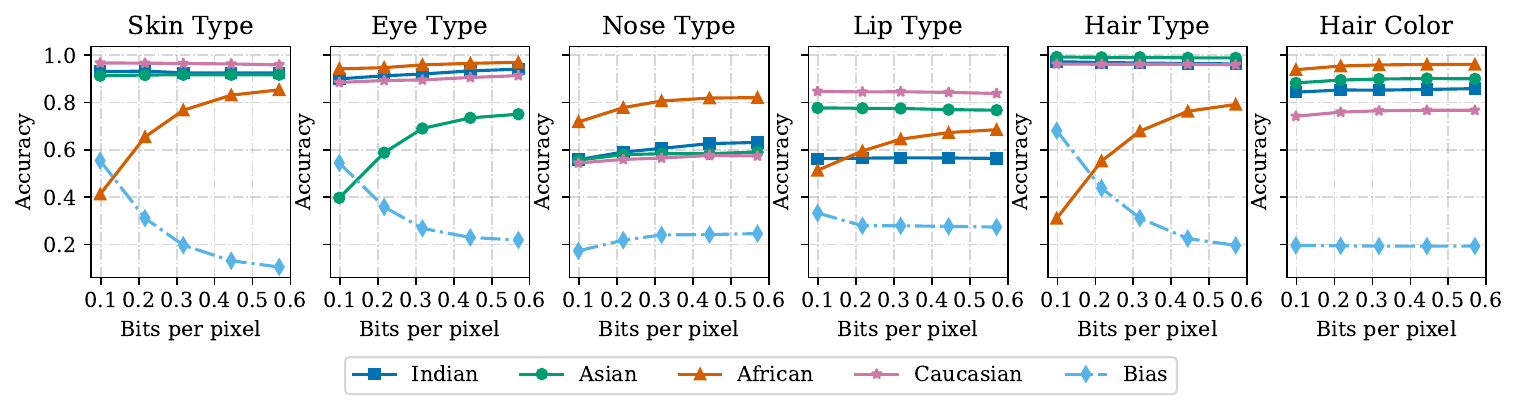}
    \caption{
    Bias in phenotype degradation for the 
 \textit{VarQRes} Model trained on CelebA
    }
\end{figure}

\begin{figure}[h!tbp]
    \includegraphics[width=\columnwidth]{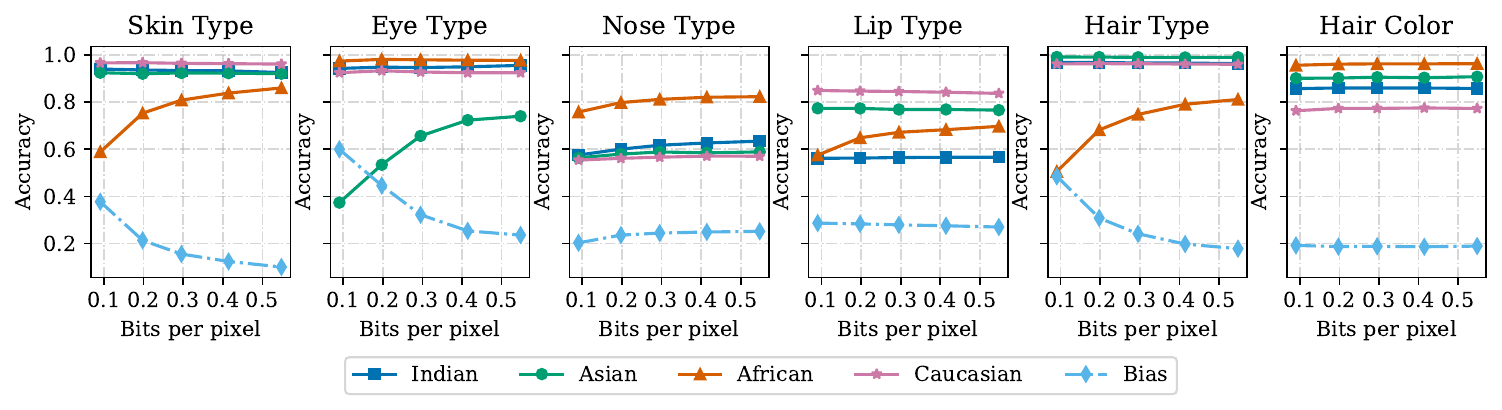}
    \caption{
    Bias in phenotype degradation for the 
 \textit{VarQRes} Model trained on FaceARG
    }
\end{figure}

\begin{figure}[h!tbp]
    \includegraphics[width=\columnwidth]{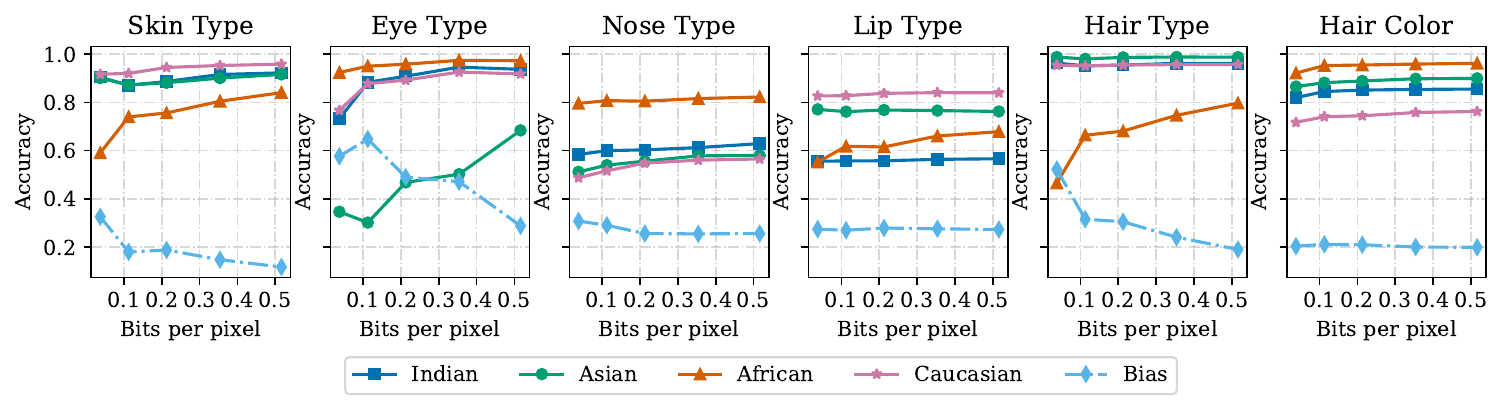}
    \caption{
    Bias in phenotype degradation for the \textit{HiFiC} Model trained on CelebA
    }
\end{figure}

\begin{figure}[h!tbp]
    \includegraphics[width=\columnwidth]{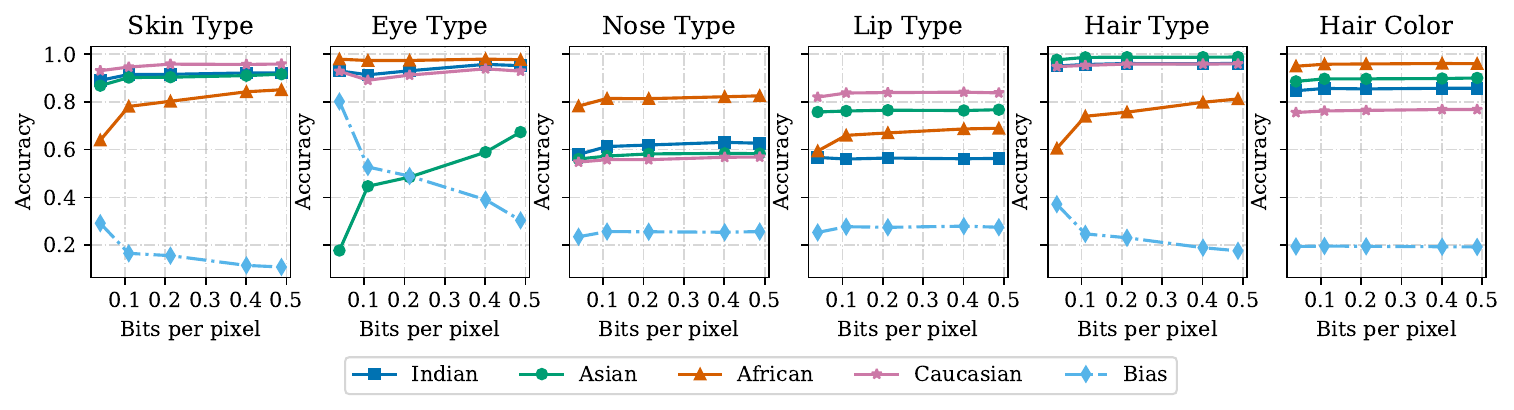}
    \caption{
    Bias in phenotype degradation for the \textit{HiFiC} Model trained on FaceARG
    }
\end{figure}

\begin{figure}[h!tbp]
    \includegraphics[width=\columnwidth]{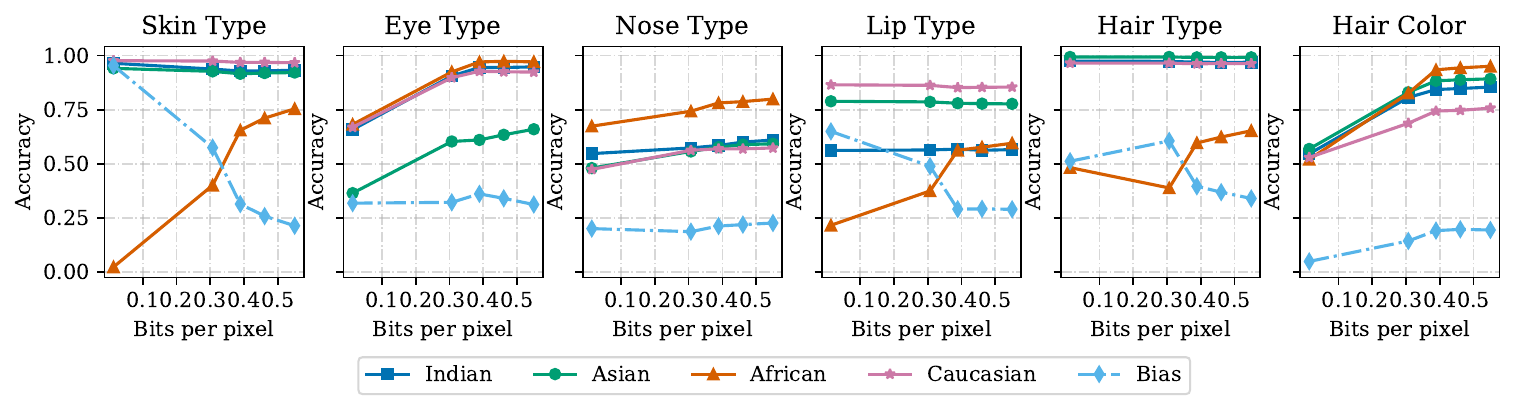}
    \caption{
    Bias in phenotype degradation for the \textit{CDC} Model trained on CelebA
    }
\end{figure}

\begin{figure}[h!tbp]
    \includegraphics[width=\columnwidth]{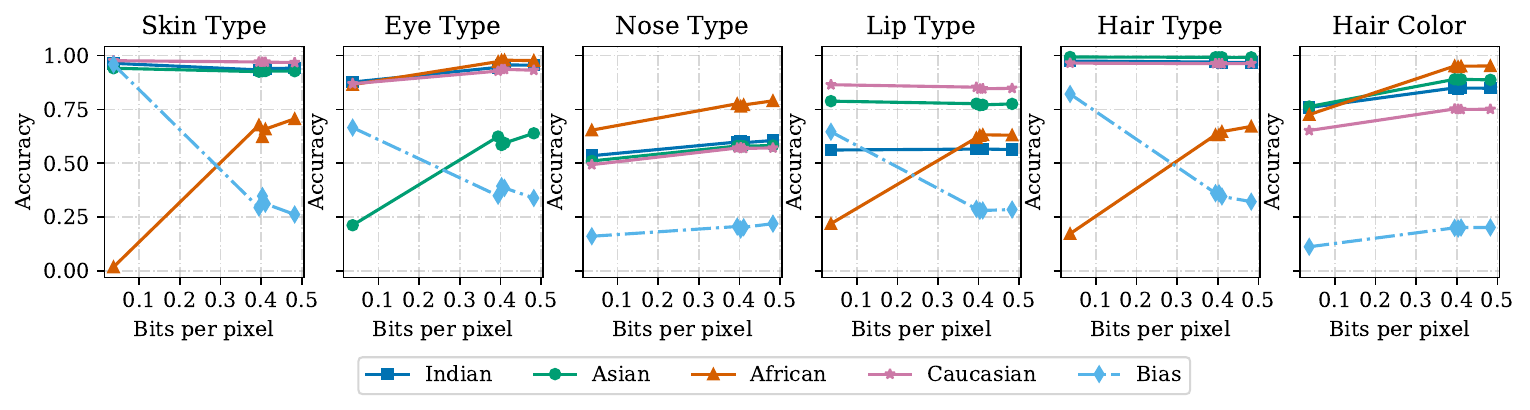}
    \caption{
    Bias in phenotype degradation for the \textit{CDC} Model trained on FaceARG
    }
\end{figure}

\begin{figure}[h!tbp]
    \includegraphics[width=\columnwidth]{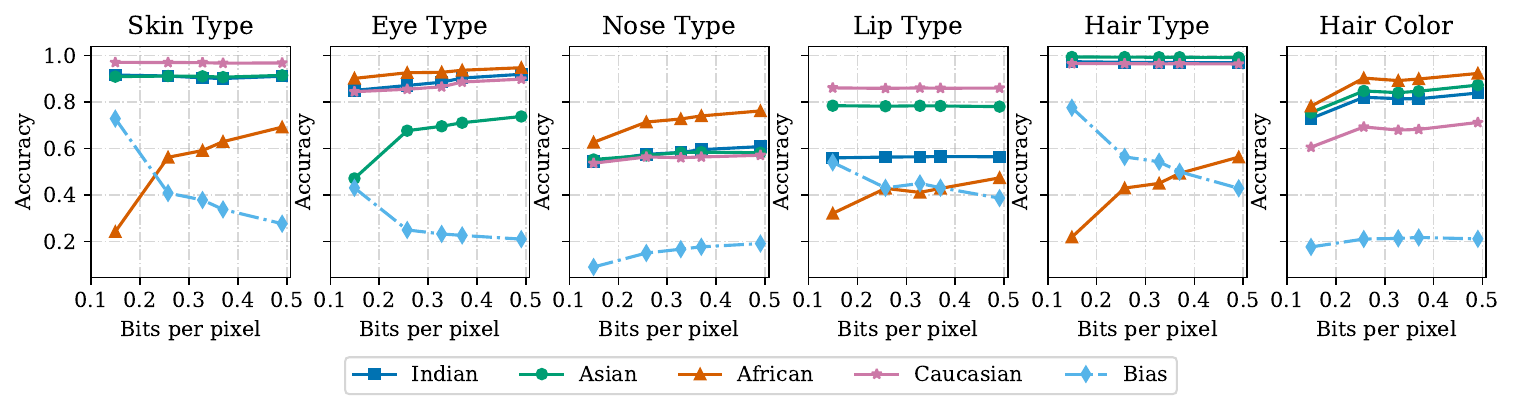}
    \caption{
    Bias in phenotype degradation for the \textit{CDC-L2} Model trained on CelebA
    }
\end{figure}

\begin{figure}[h!tbp]
    \includegraphics[width=\columnwidth]{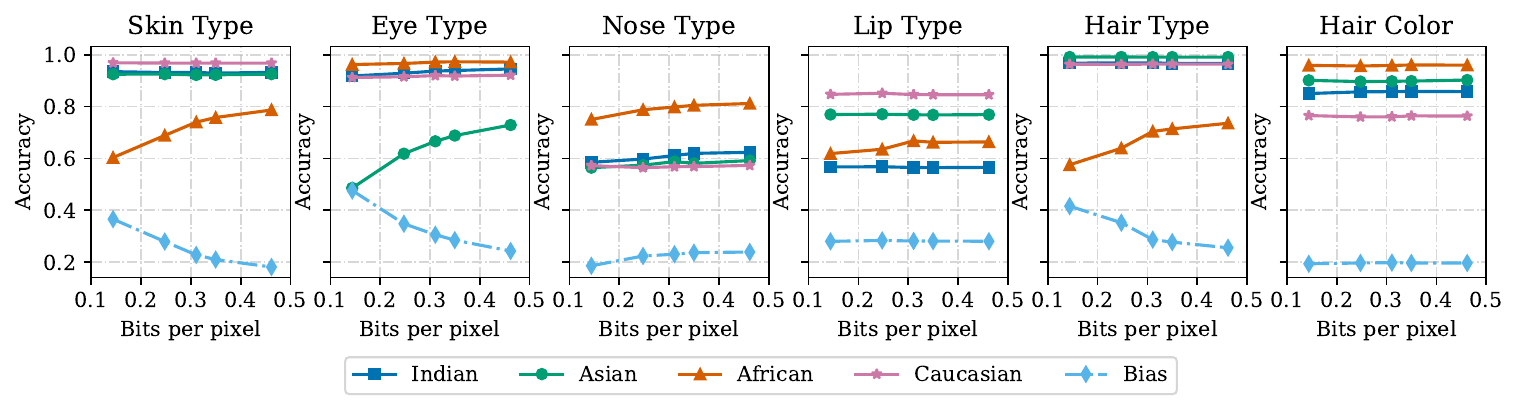}
    \caption{
    Bias in phenotype degradation for the \textit{CDC-L2} Model trained on FaceARG
    }
\end{figure}

\begin{figure}[h!tbp]
    \includegraphics[width=\columnwidth]{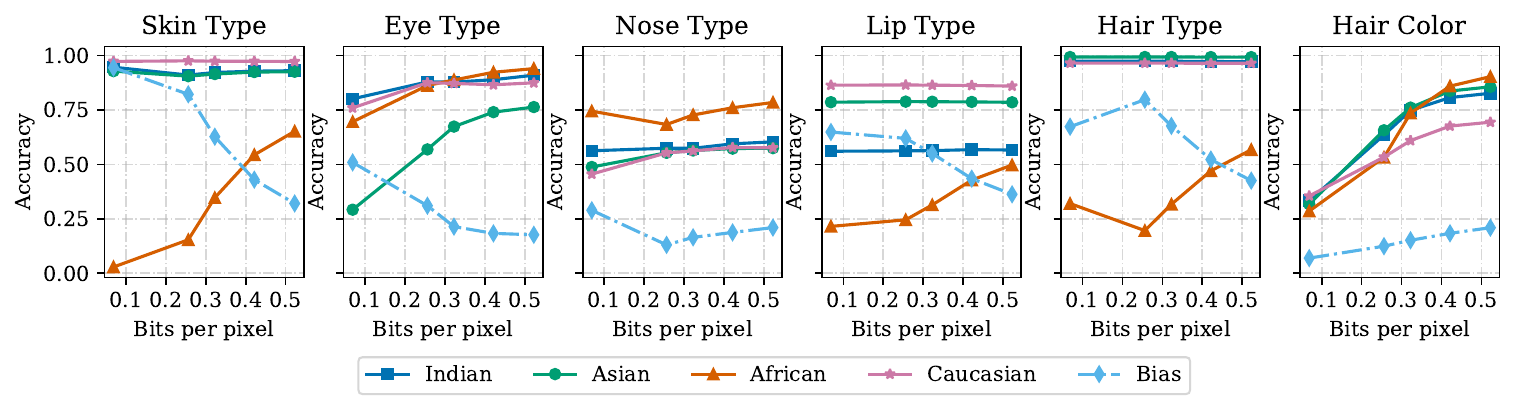}
    \caption{
    Bias in phenotype degradation for the \textit{CDC-LPIPS} Model trained on CelebA
    }
\end{figure}

\begin{figure}[h!tbp]
    \includegraphics[width=\columnwidth]{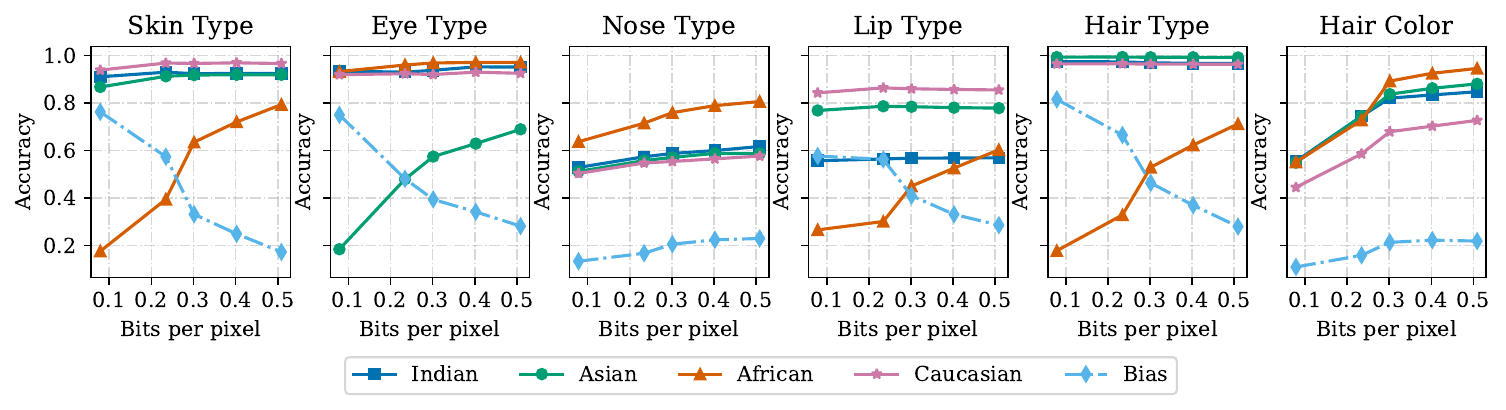}
    \caption{
    Bias in phenotype degradation for the \textit{CDC-LPIPS} Model trained on FaceARG
    }
\end{figure}

\clearpage
\newpage
\section{Bias across architectures}
\begin{figure}[h!] 
    \centering
    \includegraphics[width=0.7\textwidth]{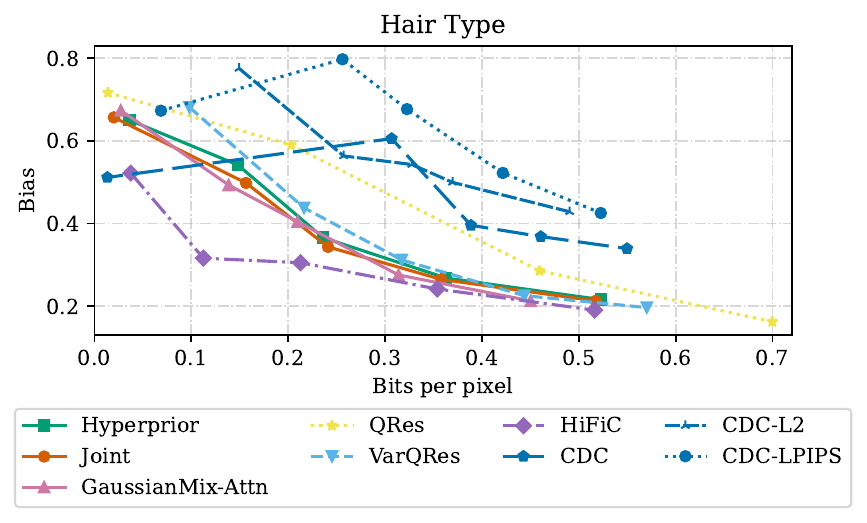}
    \caption{The trend in bias of \textit{hair type} is similar to that of the \textit{skin type}.} \label{appendix:hair_type_bias_across_models}
\end{figure}

\begin{figure}[h!] 
    \centering
    \includegraphics[width=0.8\textwidth]{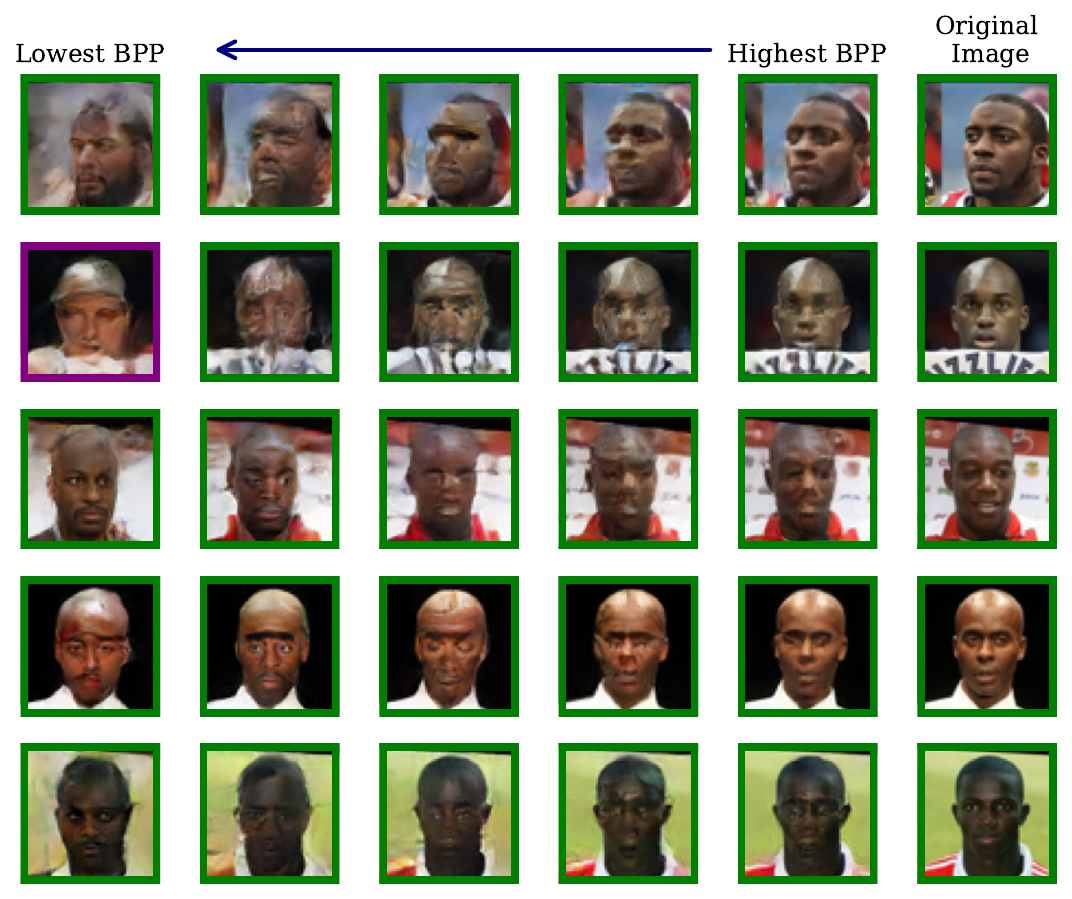}
    \caption{\textit{HiFiC} preserves \textit{skin type} well. However, it introduces extra image details.} \label{appendix:hific}
\end{figure}

\clearpage
\newpage
\section{Training with a Balanced Dataset}\label{appendix:balanced}
In Figure~\ref{fig:all_dataset_comparisons} we present the impact of using a balanced training set FaceARG on racial bias in phenotype degradation.

\begin{figure}[h!tbp]
    \centering
    \subfigure[\textit{Hyperprior} ]{\includegraphics[width=0.35\columnwidth]{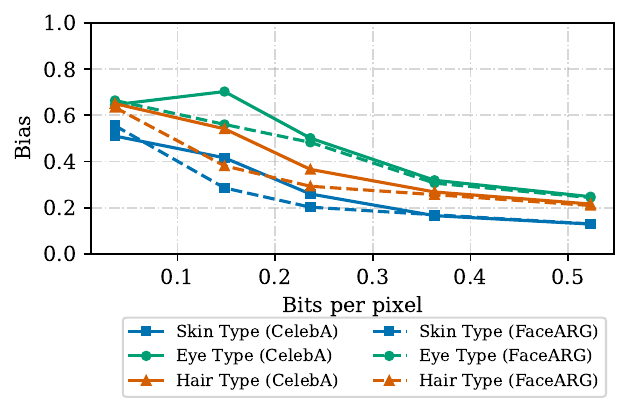}} 
    \subfigure[\textit{GaussianMix-Attn}]{\includegraphics[width=0.35\columnwidth]{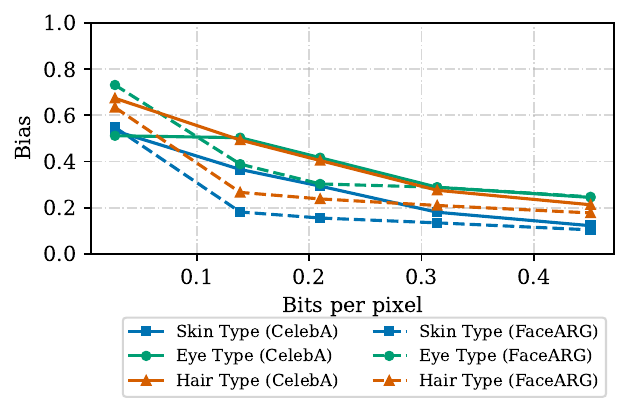}} 
    \subfigure[\textit{QRes}]{\includegraphics[width=0.35\columnwidth]{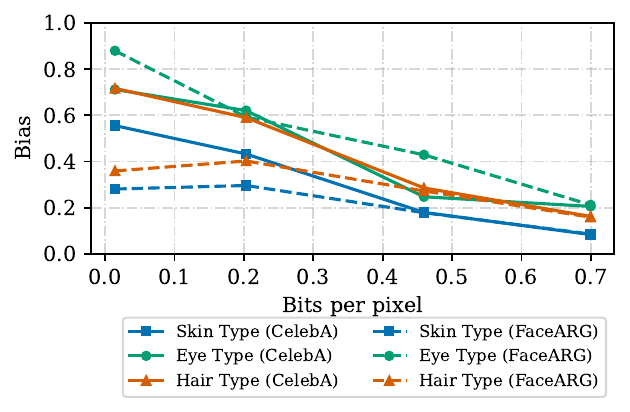}} 
    \subfigure[\textit{VarQRes} ]{\includegraphics[width=0.35\columnwidth]{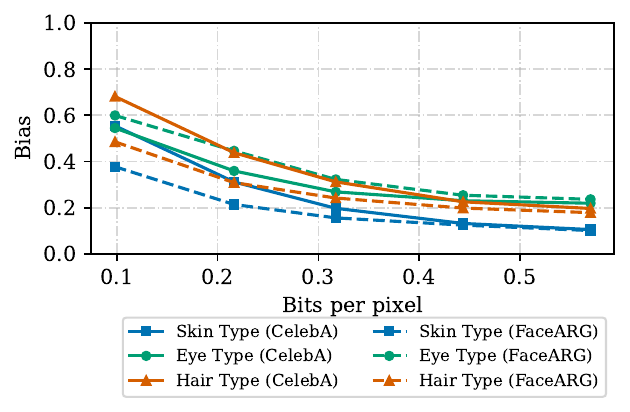}} 
    \subfigure[\textit{HiFiC}]{\includegraphics[width=0.35\columnwidth]{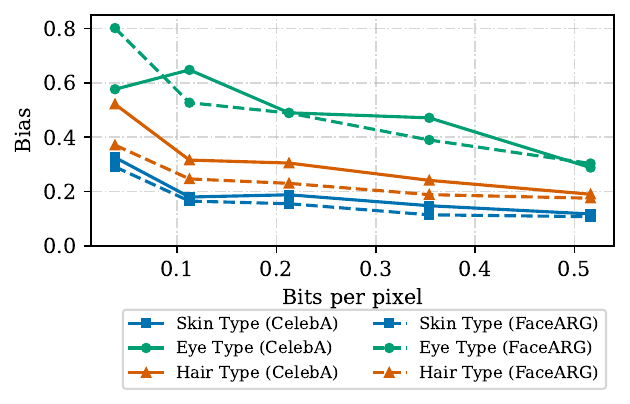}} 
    \subfigure[\textit{CDC}]{\includegraphics[width=0.35\columnwidth]{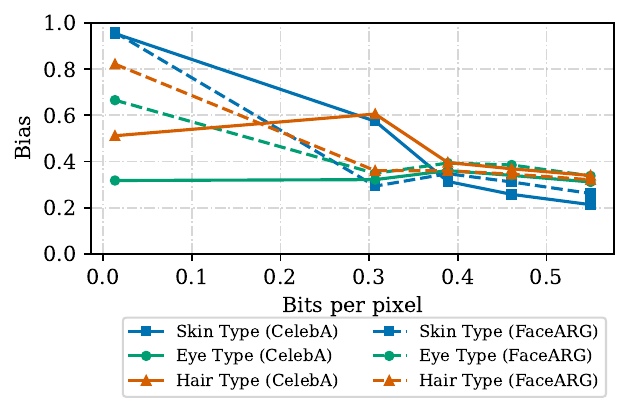}} 
    \subfigure[\textit{CDC-L2}]{\includegraphics[width=0.35\columnwidth]{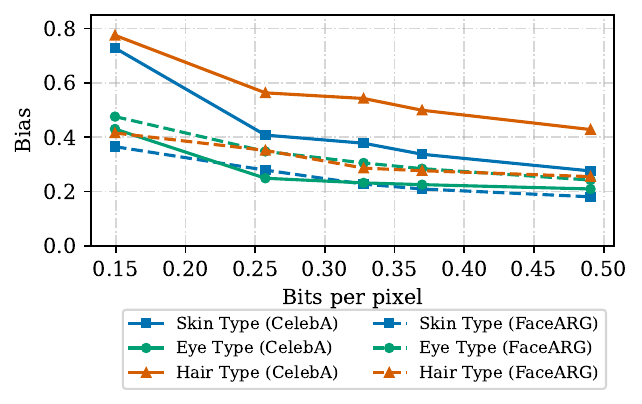}} 
    \subfigure[\textit{CDC-LPIPS}]{\includegraphics[width=0.35\columnwidth]{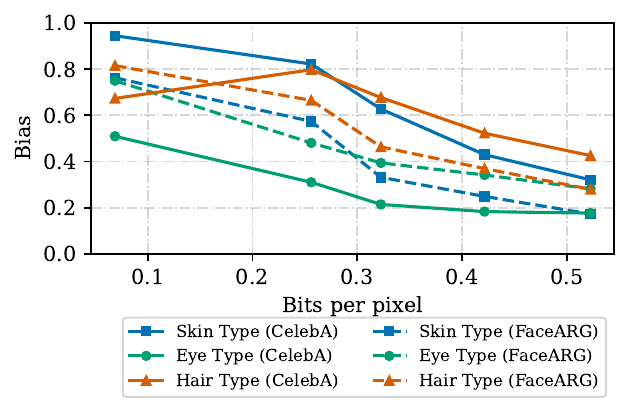}} 
    \caption{Impact on phenotype degradation bias of racially balanced or imbalanced datasets}
    
\end{figure}\label{fig:all_dataset_comparisons}

\newpage
\section{Bias-Realism Relationship} \label{appendix:trade1}
\begin{figure}[h!tbp]
    \centering
    \subfigure[Skin Type]{\includegraphics[width=0.45\columnwidth]{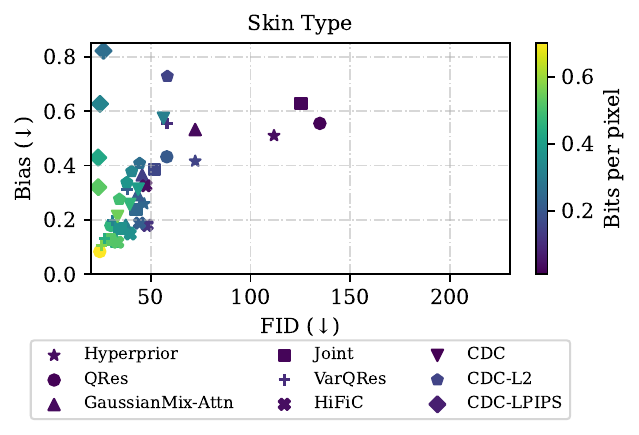}} 
    \subfigure[Lip Type ]{\includegraphics[width=0.45\columnwidth]{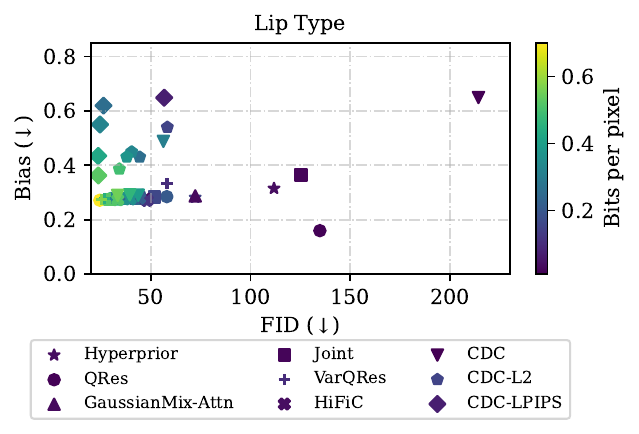}} 
    \subfigure[Nose Type]{\includegraphics[width=0.45\columnwidth]{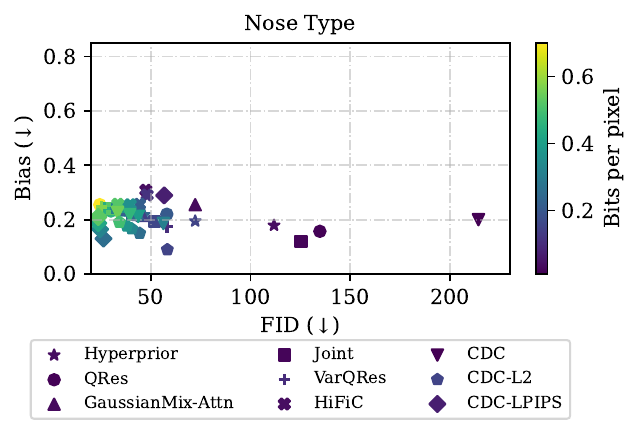}} 
    \subfigure[Eye Type]{\includegraphics[width=0.45\columnwidth]{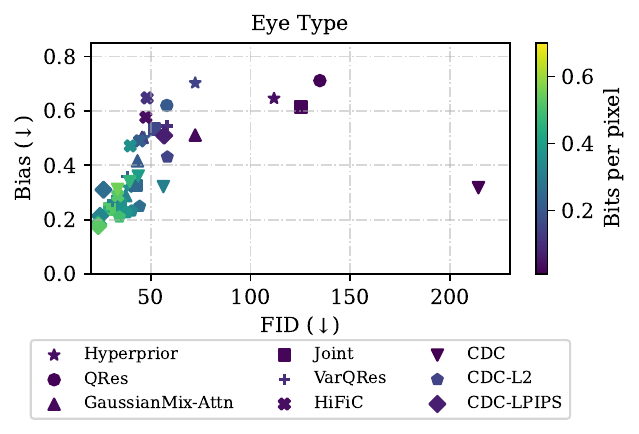}} 
    \subfigure[Hair Type]{\includegraphics[width=0.45\columnwidth]{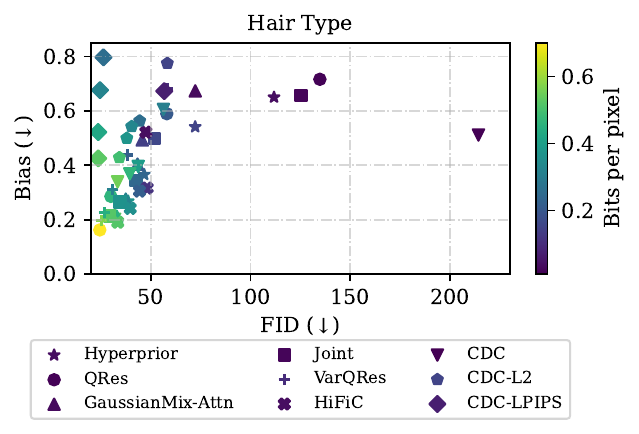}} 
    \subfigure[Hair Color]{\includegraphics[width=0.45\columnwidth]{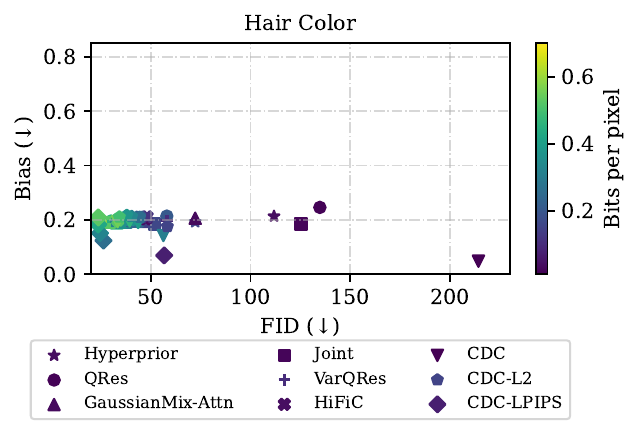}} 
    \caption{Bias-realism relationship for models trained on CelebA}
    \label{fig:all_fid_figures_celeba}
\end{figure}

\begin{figure}[t] 
    \centering
    \subfigure[Skin Type]{\includegraphics[width=0.45\columnwidth]{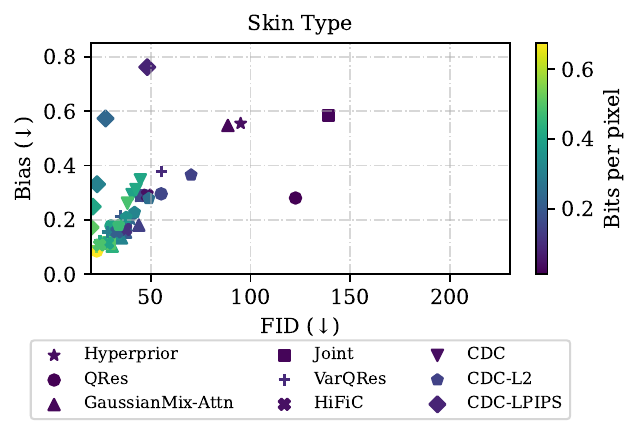}} 
    \subfigure[Lip Type ]{\includegraphics[width=0.45\columnwidth]{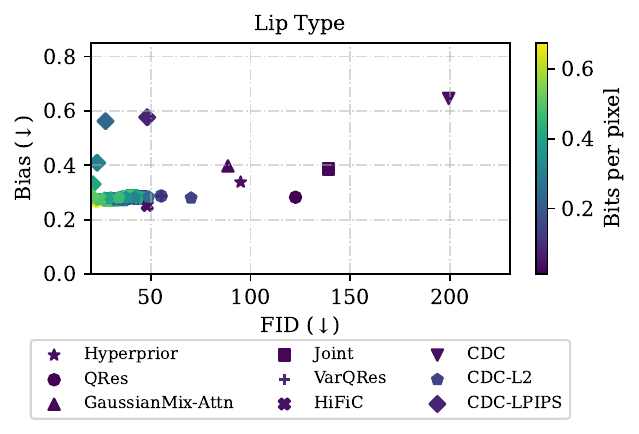}} 
    \subfigure[Nose Type]{\includegraphics[width=0.45\columnwidth]{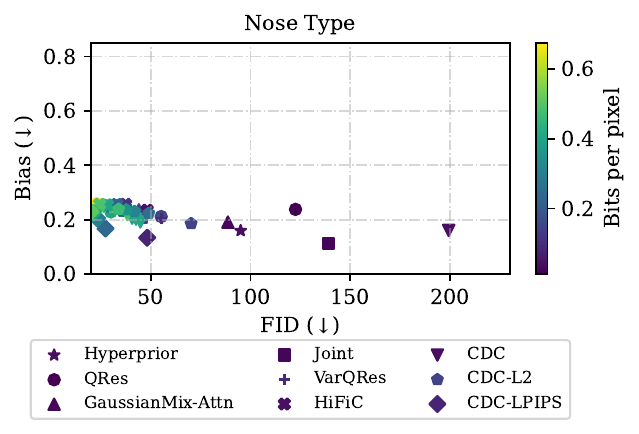}} 
    \subfigure[Eye Type]{\includegraphics[width=0.45\columnwidth]{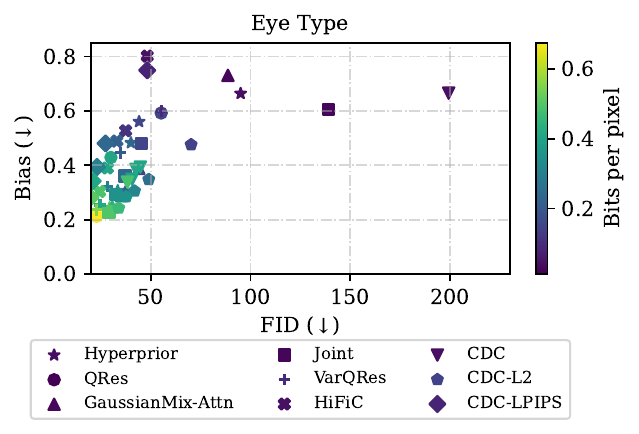}} 
    \subfigure[Hair Type]{\includegraphics[width=0.45\columnwidth]{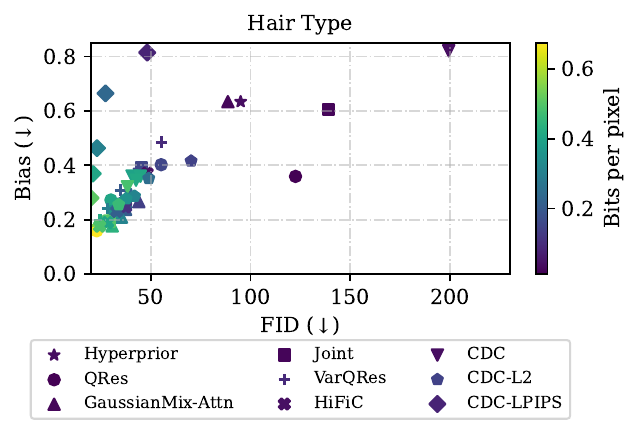}} 
    \subfigure[Hair Color]{\includegraphics[width=0.45\columnwidth]{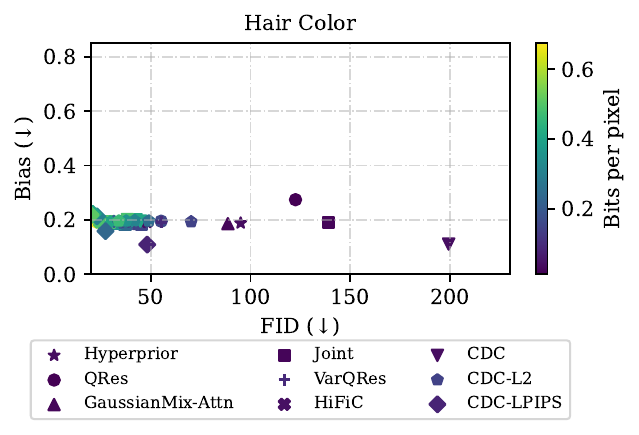}} 
    \caption{Bias-realism relationship for models trained on FaceARG}
    \label{fig:all_fid_figures_facearg}
\end{figure}

In Figure~\ref{fig:all_fid_figures_celeba} and Figure~\ref{fig:all_fid_figures_facearg} we present FID vs bias figures for all the phenotypes.

\clearpage
\newpage
\section{Training with African-only images}\label{appendix:african_only}

\begin{figure}[h!tbp]
    \centering
    \subfigure[\textit{Hyperprior}]{\includegraphics[width=0.45\columnwidth]{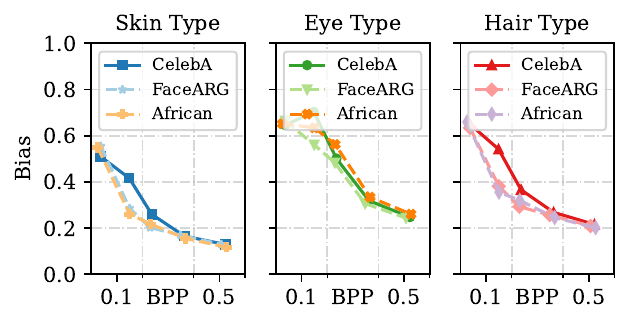}} 
    \subfigure[\textit{Joint}]{\includegraphics[width=0.45\columnwidth]{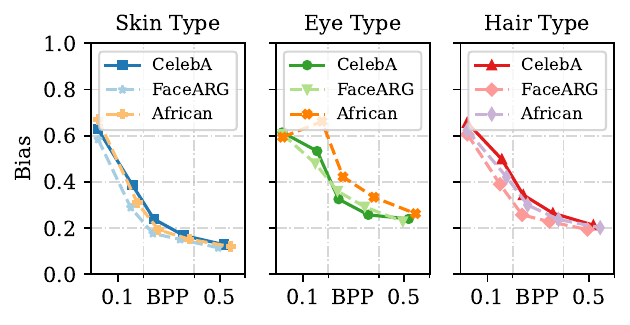}} 
    \subfigure[\textit{GaussianMix-Attn}]{\includegraphics[width=0.45\columnwidth]{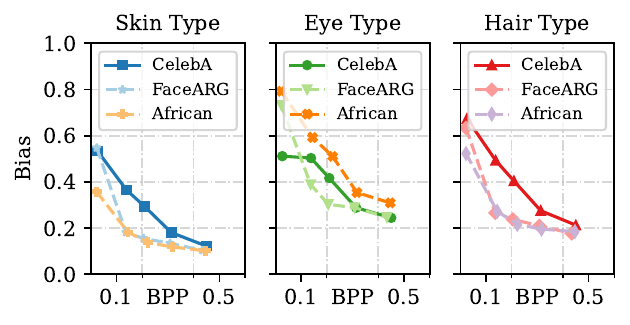}} 
    \subfigure[\textit{QRes}]{\includegraphics[width=0.45\columnwidth]{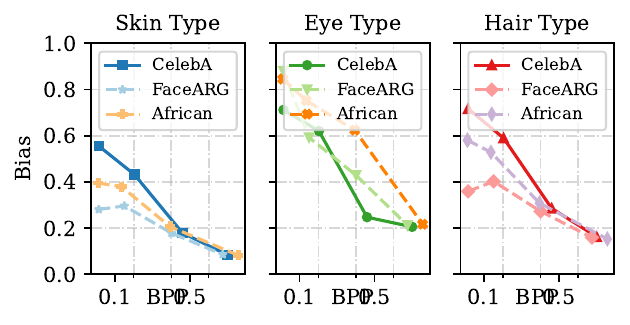}} 
    \subfigure[\textit{VarQRes}]{\includegraphics[width=0.45\columnwidth]{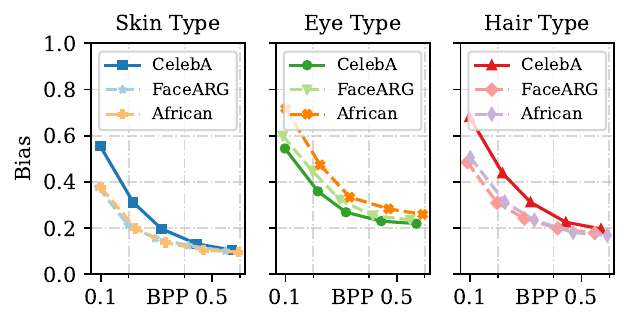}} 
    \caption{Using the african-only subset from FaceARG helps reduce bias in one model (\textit{GaussianMix-Attn}), but doesn't have significant impact on other models, suggesting bias towards African racial group is not completely removed by training with exclusively African facial images. }
\end{figure}\label{fig:african_only_training}

\begin{figure}[h!tbp]
    \centering
    \subfigure[]{\includegraphics[width=0.45\columnwidth]{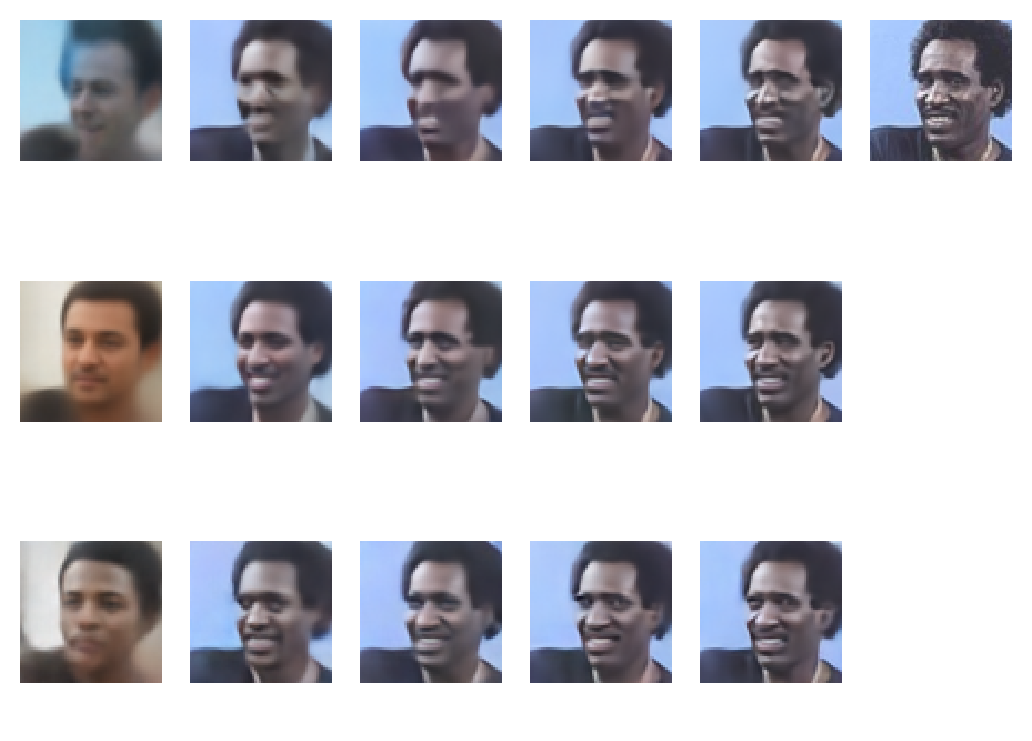}} 
    \subfigure[]{\includegraphics[width=0.45\columnwidth]{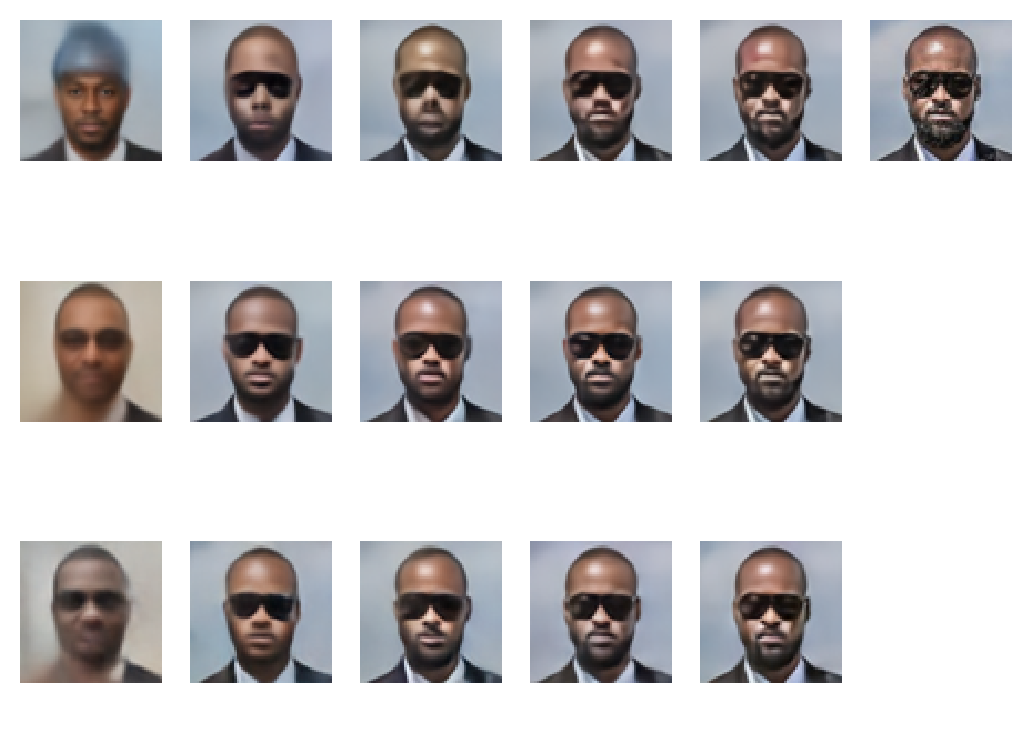}} 
    \subfigure[]{\includegraphics[width=0.45\columnwidth]{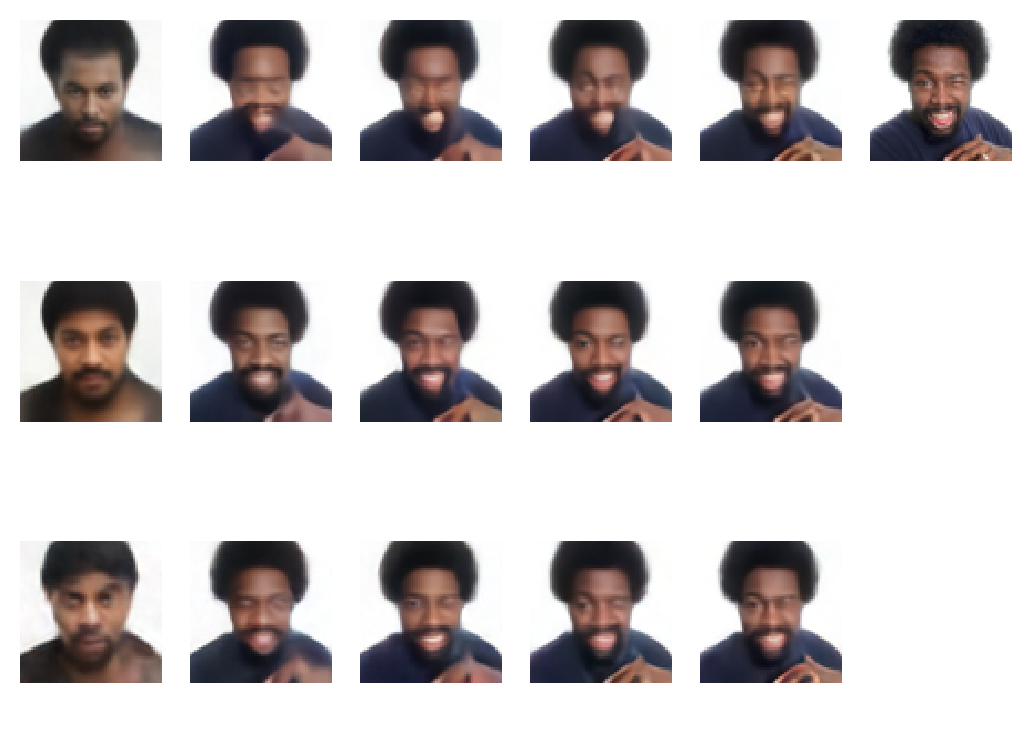}} 
    \subfigure[]{\includegraphics[width=0.45\columnwidth]{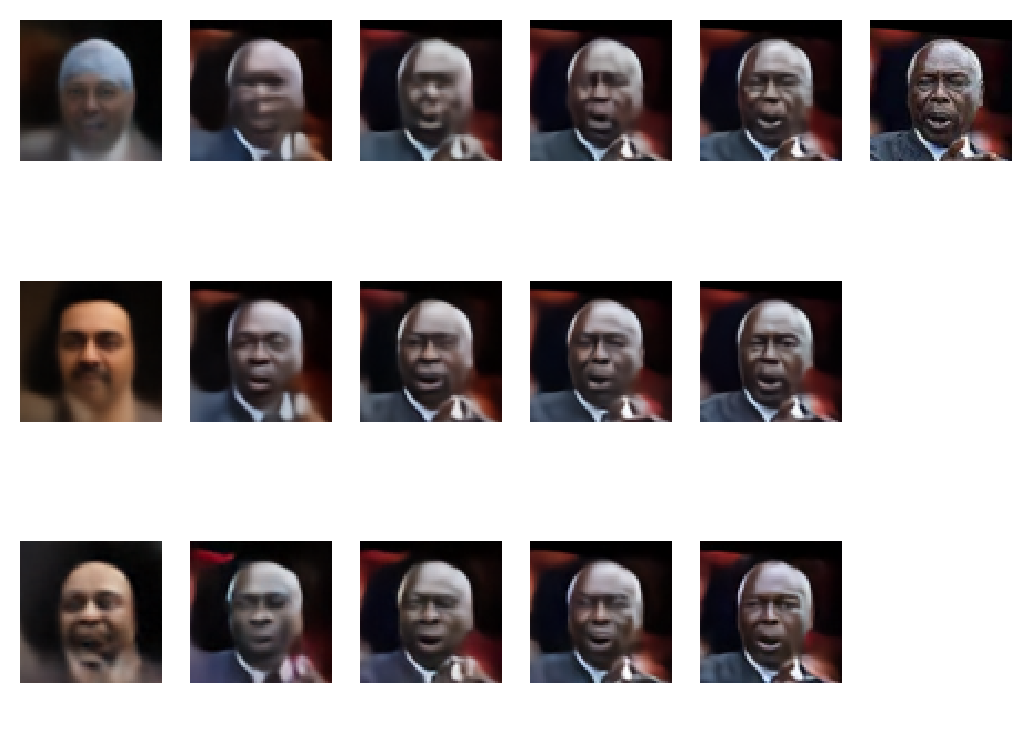}} 
    \caption{In each subfigure, the reconstructions are from models trained with CelebA (top), FaceARG (middle), and African subset from FaceARG (bottom). The bitrate reduces from right to left, with rightmost image the original image. (a) and (b): Examples of training with african only reduces skin type bias. (c) and (d): Examples of skin type bias still exists after training with african only images.}
\end{figure}\label{fig:african_visual}

\clearpage
\newpage
\section{Details on Human Annotations}\label{appendix:annotation}
\paragraph{The task}
To validate our findings with the phenotype classifier, we additionally acquired human annotations of clean and decoded images from the model. 
Each participant is asked to look at a sequence of images and classify each image into one of six classes based on Fitzpatrick skin type.
\paragraph{Participant recruitment}
We obtained the human annotations on Prolific, a well-adopted platform for online studies. Participants are recruited using Prolific’s standard sampling procedure (distributing the study to all available participants). Pre-screening procedure is included to only recruit participants that are fluent in English. As a result, the recruiting participants reflects the general distribution of Prolific workers. From the information logged with Prolific, we find that there were 51.7\% female 47.5\% male, and 52\% white, 28\% black, 30\% others. Participant’s explicit consent was acquired before proceeding to the task.
\paragraph{Experiment design}
The study includes two tasks: one for labeling clean images and another for compressed images. Sixty pairs of clean and compressed images are randomly sampled from the RFW test set, with clean images in one task and compressed images in the other. For each task, 100 participants are recruited. For this study, we only focus on images under the African racial group as annotated in the RFW dataset.

Before the task starts, the participant is provided with a training session, where randomly sampled images from all the six Fitzpatrick skin types are provided for the participant to study. After the training section, there is a calibration session to test the participant’s understanding the skin type scale, where the participant is asked to classify images as in the task, but will be provided the ground truth label afterwards to calibrate their understanding of the scale.

\end{document}